\documentclass[fleqn,10pt]{wlscirep}
\usepackage[utf8]{inputenc}
\usepackage[T1]{fontenc}
\usepackage{adjustbox,multirow}
\usepackage{lineno}
\usepackage{subcaption,graphicx}
\title{AI Assisted Cervical Cancer Screening for Cytology Samples in Developing Countries}

\author[1,*]{Love Panta}
\author[1]{Suraj Prasai}
\author[2]{Karishma Malla Vaidya}
\author[1]{Shyam Shrestha}
\author[1]{Suresh Manandhar}
\affil[1]{Wiseyak www.wiseyak.com}
\affil[2]{Paropakar Prasuti Griha Maternity Hospital, Nepal}
\affil[*]{love.panta@wiseyak.com}

\begin{abstract}
Cervical cancer remains a significant health challenge, with high incidence and mortality rates, particularly in transitioning countries. Conventional Liquid-Based Cytology (LBC) is a labor-intensive process, requires expert pathologists and is highly prone to errors, highlighting the need for more efficient screening methods. This paper introduces an innovative approach that integrates low-cost biological microscopes with our simple and efficient AI algorithms for automated whole-slide analysis. Our system uses a motorized microscope to capture cytology images, which are then processed through an AI pipeline involving image stitching, cell segmentation, and classification. We utilize the lightweight UNet-based model involving human-in-the-loop approach to train our segmentation model with minimal ROIs. CvT-based classification model, trained on the SIPaKMeD dataset, accurately categorizes five cell types. Our framework offers enhanced accuracy and efficiency in cervical cancer screening compared to various state-of-art methods, as demonstrated by different evaluation metrics.
\end{abstract}
\begin{document}

\flushbottom
\maketitle
%
%
\thispagestyle{empty}

\section{Introduction}
Cervical cancer is the fourth most common cancer globally in terms of both incidence and mortality among women \cite{bray2024global}. The Global Cancer Statistics 2022 report by IARC and ACS indicates that approximately 1,800 women were diagnosed with cervical cancer, with nearly 1,000 deaths. The rates remain disproportionately high in transitioning countries compared to transitioned ones, with incidence rates of 19.3 per 100,000 versus 12.1 per 100,000, and mortality rates of 12.4 per 100,000 versus 4.8 per 100,000, respectively \cite{bray2024global}. Several factors, such as sexually transmitted diseases (STDs), high parity, cigarette smoking, and overuse of oral contraceptives, can contribute to the development of cervical cancer in conjunction with HPV types \cite{okunade2020human}. Early preventive measures, including HPV testing, cervical cytology screening, and vaccination, can significantly reduce the risk of cervical cancer.
\par
Liquid Based Cytology(LBC) is an effective early screening method which involves manual inspection of the abnormal cells under electron microscopes by the expert pathologist. The process is labor-intensive and highly prone to error. So, an effective solution is needed to develop the system which could increase the screening frequency to reduce the mortality rate caused by cervical cancer. Moreover, the solution should also be efficient and affordable to low-income countries. The best approach is to use low-cost microscopes together with advanced AI algorithms to automate the screening process. \cite{shen2023cost} also assessed the cost-effectiveness of AI-assisted LBC testing for 6 different screening frequencies, concluding the test conducted every 5 years to be the most appropriate solution. But these results depend upon many factors such as screening population size, variation in screening sensitivity and specificity, and  HPV test cost. \cite{yang2024clinical} also performed the clinical evaluation of AI-based cytology alone and in conjunction with other methods. They found the approach superior to the recommended LBC + HPV cotesting strategy for triaging patients. 
\par
With the increasing integration of Artificial Intelligence (AI) across various sectors, its application in healthcare diagnosis has become prevalent due to the surge in medical data availability. Numerous deep learning-based methods have been introduced in recent years, addressing tasks such as medical image segmentation, disease classification, and 3D bone reconstruction. As the world becomes more digitized, vast amounts of patient data are now stored electronically in the form of Electronic Health Records (EHR), facilitating the deployment of AI-driven applications. Cervical cytology slides, similarly, require digitization to be converted into a usable format. In this context, mobile devices equipped with low-cost biological microscopes serve as a practical solution for scanning the entire surface area of targeted samples. This approach not only enhances data accessibility and storage but also mitigates risks associated with data loss or equipment failure. Once captured, the imaging data can be analyzed—either manually or through automated systems—to extract vital information, accelerating and improving the efficiency of cervical cancer screening.
\par
In this study, we propose a novel technique for whole slide analysis with an integration of low-cost biological microscopes along with an effective AI system. The microscope is automated with the use of a motorized system along the orthogonal axis for planar motion which enables capturing of target cytology videos with a digital camera mounted on its neck. On the software side, an inference system is developed that post-process the captured videos with the aid of sequential AI models to provide final medical reports. The image-stitching pipeline is introduced to build the panoramic view of cytology images. It is followed by the cell-segmentation module that perform segmentation at cell level. The training process of this segmentation method is inspired from existing pre-trained models which utilizes a diverse set of cell images annotated with unique labels for computing accurate flow fields representing pixel movement and distance to object centers. The approach is differ from instance segmentation in which we directly predict masks for each cell instance. Later, these segmented cells are passed through separate classification models to distinguish between five different cell classes. The classification model utilizes the robust and computationally efficient CvT model for training on SIPaKMeD dataset. The overall framework is evaluated on our collected cytology images to show its generalizability.
\par
In summary, our contributions are three folds:
\begin{itemize}
\item We propose a novel and comprehensive approach for analyzing cervical cytology samples with the integration of low cost microscopes and our efficient AI pipeline.
\item Lightweight and effective segmentation module called cellpose2.0 is finetuned and evaluated on cervical cell images. We also built segmentation dataset called CYTOCERVIX collected from a variety of cytology slide samples with different cell size, object density, variation in colors, texture to improve generalization in unseen test data.
\item For the first time, the Convolution Vision transformer(CvT) based model is utilized for cervical cell classification tasks outperforming various recent SOTA methods on 5-class SIPaKMeD dataset.
\end{itemize}

\section{Literature Review}
The review papers by \cite{hou2022artificial,allahqoli2022diagnosis,mustafa2023cervical} discussed the potential applications of AI for the early screening of cervical cancers, highlighting how it reduces the labor-intensive process and the high error rate of manual observation. They are mainly based on dividing the overall pipeline into two main stages i.e segmentation and classification enhancing their applicability in automatic analysis of pap smear for improving the efficiency of screening. In the survey paper \cite{hu2023state}, the authors review the Computer Aided Diagnosis(CAD) techniques using Whole slide Image Analysis for diseases, such as, gastric cancer, lung cancer, breast cancer and colon cancer. The paper discussed the limitations of traditional ANNs for large-sized images, ultimately leading to the use of DNN and Transformer-based architectures for WSI CAD. Jiang et al. first performed the specialized review of cervical cytology for automatic WSI analysis \cite{jiang2023systematic}. The paper also provides the biomedical knowledge needed for detailed understanding of its applicability in related fields. Moreover, it investigates various approaches for applying object detection algorithms for cytology screening, its challenges and future directions. Rahaman et al. \cite{rahaman2020survey} studied the state-of-the-art deep learning (DL) methods for cervical cytopathology images using the Herlev dataset \cite{jantzen2005pap}, and the ISBI challenge dataset as popular benchmarks for image segmentation. Their review also highlights extensive works done in the ensemble methods and the combination of DL and transfer learning for cell classification. They also show the potential application of mobile phone microscopy for automatic slide analysis.
\subsection{Cell Segmentation}
Segmentation of cells is a challenging task that requires effective modeling of complex cell structures from multiple heterogeneous sources, accurate separation of overlapping cells, and so on. So, several reviews have been proposed till date that studied the most recent and commonly used AI techniques for robust cell segmentation \cite{vicar2019cell,rahaman2020survey,aswath2023segmentation}. StarDist \cite{schmidt2018cell} is one such technique which utilizes the lightweight U-Net \cite{ronneberger2015u} segmentation model for wide-range of crowded cells. The method is based on computing Star-convex polygon distances(a radial distance originating from each pixel position to the boundary of each object) and object probabilities(binary masks) followed by Non-maximum suppression for representing each cell object instance. They showed competitive results in comparison to Mask-RCNN with much fewer parameters and easy training procedures. Similarly, \cite{stringer2021cellpose} firstly introduces a generalized segmentation model called Cellpose based on U-net architecture trained with an objective of predicting topological maps and probability indicating each pixel belongs to a given cell. This ensures that flow of all pixels are routed towards the center thereby allowing separation of each cell much easier. It uses a high variation of cell images with over 70,000 annotations for training purposes. They outperformed previous methods at various thresholds. Cutler et al. further combine the predicted outputs of StarDist, Cellpose and Misic \cite{panigrahi2021misic} to develop their new-bacterial cell segmentation model called OmniPose \cite{cutler2022omnipose}. On the other hand, extending the work of Cellpose, \cite{pachitariu2022cellpose} presents Cellpose2.0,  which utilized the pretrained model trained on highly diverse Cellpose dataset to perform fine tuning with only 500-1000 manual-annotated ROIs. The results are comparable with the model trained from scratch using 200,000 ROIs. Additionally, it incorporates human-on-the-loop pipeline requiring minimal human annotations ranging from 100-200 ROIs without compromising segmentation quality.  

\subsection{Cell Classification}
Most of previous approaches \cite{plissiti2018sipakmed,wang2019automatic} are based on hand-craft feature extractions involving segmentation of the region of interest, calculation of features based on shape, intensity, texture and so on, limiting their effectiveness for cell classification. \cite{basak2021cervical,win2020computer} utilize feature selection algorithms such as PCA, GWO, RF followed by a cell classifier integrating either single SVM or its ensemble(fusion of SVM, KNN and other).  Fang et al. introduces three versions of CNN architectures called DeepCELL-vX  that learns feature representations using multiple kernels of different size to classify cervical cytology images for Herlev and SIPaKMeD datasets \cite{fang2022deep}. Regarding multi-modal fusion, \cite{pramanik2022fuzzy,manna2021fuzzy} propose fuzzy based ensembles combining three CNN models for automatic detection of cervical cancer on PaP Smear datasets. First each model is trained using transfer learning methods and then ensemble approaches based on minimizing differences between observed solutions and ideal solutions are added. Finally, the minimum argument of defuzzified values is taken as the final prediction. In context of Whole Slide Image (WSI), recently artiﬁcial intelligence cervical cancer screening (AICCS) system \cite{wang2024artificial} was created functioning at three major stages. Initially, whole slide images undergo patch-level annotation in a sliding mode-fashion. Then, the cell detection model called RetinaNet is trained to generate target ROIs which serves as input for WSI level classification methods which in turn generate all possible cytology grades: ASC-US, LSIL, ASC-H, HSIL, SCC, and AGC. The extensive experiments and prospective evaluation validates the improved performance on randomized observational trials exhibiting its potential for efficient and precise screening of cervical Cancer. In the research of instance segmentation, \cite{sompawong2019automated} first applied Mask R-CNN for automated screening of cervical cancer from pap smear histological slides. This approach detects the bounding box of each nucleus along with finding its type as normal or abnormal cells. Furthermore, they introduce a modified version of  DeepPap \cite{zhang2017deeppap} based on transfer learning using InceptionV3 to compare their performance with Mask R-CNN for single cell classification. Likewise, Graph Convolution Network is explored in the following paper \cite{shi2021cervical} for cell classification. It is based on the assumption that images are highly correlated on the latent dimension and this is totally  ignored by CNN feature learning module thus reducing their representation ability on different images. So, they construct the graph structure based on centroid features derived from clustering algorithms and assign each individual image feature to one of the clusters. The GCN then learns the relation aware context by  incorporating multiple cluster centroids which is later used to model discriminative class representation for cell classification.
\par
Several ensemble approaches combining the best CNN and ViT model are introduced in \cite{pacal2023deep}. Their best ViT-B16 outperformed EfficientNet-B6 CNN model in all classification metrics for SIPaKMeD dataset with slight improvement when using Max-Voting ensemble method. Similarly, a hybrid deep feature fusion (HDFF) technique is proposed which is based on feature extractions from multiple complementary CNN networks and fusing later for cell classification task \cite{rahaman2021deepcervix}. They have achieved higher performance in 2 class, 3 class and 5 class classification on SIPaKMeD dataset. In order to exploit both local and global context information from cell images, \cite{liu2022cvm} introduced CVM-Cervix, the fusion of Xception CNN network and tiny Deit Module without distillation tokens. CVM-Cervix first extract features from both modules and then perform simple concatenation operation on those features followed by MLP layer for 11-class cell classification task. They combined pap smear datasets obtained from \cite{rezende2021cric} and \cite{plissiti2018sipakmed} each having six and five categories respectively. However, \cite{fang2024deep} argue that this process is ineffective and results in low classification performance. So, they propose the approach that synergistically integrates both features in deep fashion to obtain richer and more discriminating representation of cell images. A novel attention feature pyramid network is  developed for automatic cervical abnormal cell detection in the following paper\cite{cao2021novel}. It makes use of multi-scale feature fusion together with attention modules to capture both local and global structure of abnormal cells. Recently, Pacal et al. introduced the modification version of MaxViT called MaxCerVixT by substituting its MBConv blocks with ConvNeXtv2 blocks and MLP blocks with GRN-based MLPs to demonstrate higher accuracy on SIPaKMeD dataset \cite{pacal2024maxcervixt}. Despite greater progression, these approaches still present difficulties in integration of additional modules, are resource intensive, dependent on positional embedding, and token size. CvT \cite{wu2021cvt} is the alternative that encompasses all the weak characteristics of above approaches and improves the performance of classification systems.

\begin{figure*}[!h]
    \includegraphics[width=\textwidth]{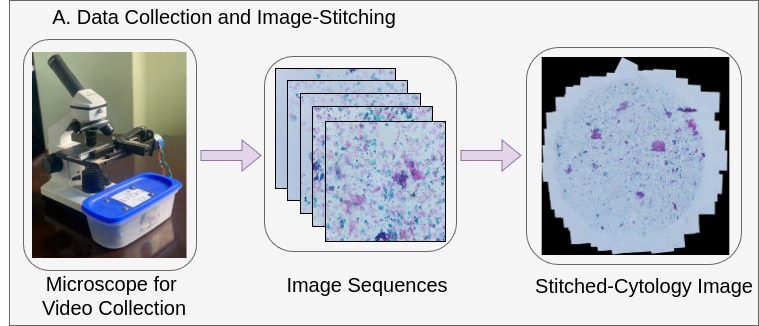}
\end{figure*}

\begin{figure*}[!h]
    \includegraphics[width=\textwidth]{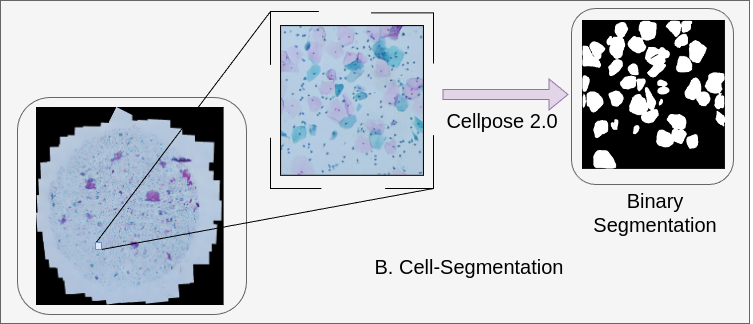}
\end{figure*} 

\begin{figure*}[!h]
   \centering
    \includegraphics[width=7.7cm]{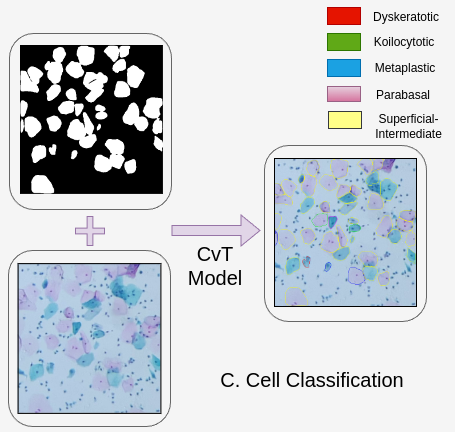}
\end{figure*} 

\section{Methodology}
We divide the whole procedure of cervical cancer diagnosis into the following four subsequent subsections. The first process is digital sample collection and its storage into the EMR system. Others involve building a panoramic view of the whole slide image, segmentation of each individual cell, and their classification as depicted in Figure (A),(B),(C). Finally, we generate the aggregate medical reports.

\subsection{Samples Collection Procedures} 
We collaborate with the various hospitals of Nepal for collecting cytology samples in Nepal. Among which Paropakar maternity and Women’s hospital is the central one facilitating most of the Cervical Cancer Diagnosis using LBC. The total of 10 cytology slide samples are collected for creating a cytology dataset named CYTOCERVIX. Those samples involve a laborious effort of scanning and analyzing anomalies using expensive electron microscopes by the highly skilled health professionals thereby prolonging the diagnosis time. To ease the process by reducing data collections time, low cost digital microscopes are integrated with mobile devices. We use the compound monocular biological microscope and replace its adjustment handle located in the mechanical stage with a suitable dc motor with an optimum speed of 16 rpm along both X-Y Axis. The wide-field Eyepiece and objective lens of the microscope are adjusted to 20X and 4X respectively with a comprehensive magnification of 80X. We use an Iphone 11 pro max as a mobile device due to its high resolution of 4K and frame rate of 60 FPS. Considering the maximum sample size of the slide, we programmed the pre planned route to scan the entire section taking an average time of 4 minutes. The recorded videos are then fetched into the next stage for further processing.

\subsection{Image-stitching Pipeline}
The image-stitching process is highly dependent on both quantity and quality of images. We must ensure that each extracted frame is overlapped as well. Stitching all the frames of video at 60 fps is also inefficient. So, we collect images at every 50 frames depending on the speed of the motor to ensure at least 25 percent overlap between each consecutive images. We used the popular opencv based image-stitching library which uses the invariant features for automatic image-stitching. The images which have no connections with any other images are automatically removed as noisy images using RANSAC and probabilistic models \cite{brown2007automatic}. In our case, all the extracted images with no cells are regarded as noisy images. We have used a sift feature detector with a confidence threshold of 0.2 for maximum matching.

\subsection{Cell Segmentation}
\begin{figure*}[!h]
    \includegraphics[width=\textwidth]{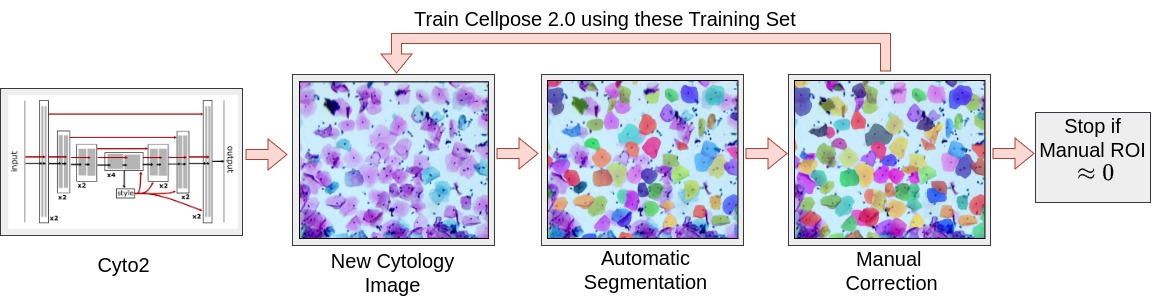}
    \caption{Human-on-the-loop approach for training our segmentation model \cite{pachitariu2022cellpose}}
    \label{segmentation_pipeline}
\end{figure*}

We utilize one of the pretrained segmentation models from cellpose2.0 called ‘cyto2’ for implementing human-on-the-loop pipeline. First of all, ‘cyto2’ is called for predicting segmentation masks on our targeted cell images. Then, the human-on-the-loop approach is used to correct any mistakes during prediction by manually drawing ROI on missing or incorrectly segmented cells as shown in Figure \ref{segmentation_pipeline}. This serves as a ground truth annotation for further refining our pretrained models. We iterate the procedures for a few other images until satisfied with segmentation accuracy. Here, we discard what each cell type is and only focus on the binary segmentation for evaluating our segmentation model. Later, each segmented cell is passed through a classification model to distinguish between different cell types.

\subsection{Cell Classification}
We have used SIPaKMeD \cite{plissiti2018sipakmed} publicly available cytology images for training our classification model. The dataset contains 4096 isolated pap smear cells annotated into five classes i.e (superficial-intermediate, parabasal) as normal, (koilocytes, dyskeratotic) as abnormal and metaplastic as benign. We split the whole image into two set using 5-fold validation strategy. Prior to training, all necessary preprocessing and transformations are done to ensure good performance on validation dataset. From an architecture perspective, we used the Convolutional Vision Transformer \cite{wu2021cvt} as shown in Figure \ref{CVT} for the training classification model on the SIPaKMeD dataset. 

\subsubsection{Convolutional Vision Transformer}
Convolutional Vision Transformer(CvT) \cite{wu2021cvt} is a computationally robust and memory efficient neural network architecture incorporating Convolution in Transformer network thereby gaining the benefit of both for image recognition tasks. Additionally, it doesn’t require any positional embedding to input tokens which in our case is independent of the resolution of cell images. We opted for this architecture due to its higher accuracy and its ability to transfer into various downstream tasks like cell classification.

\begin{figure}[!h]
    \centering
    \includegraphics[width=7cm]{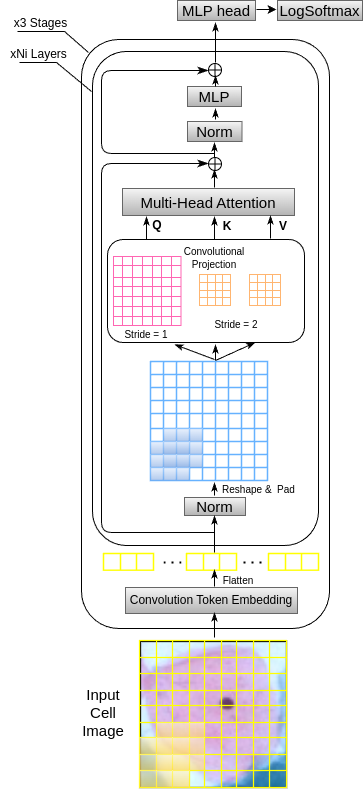}
    \caption{Convolutional Vision Transformer pipeline \cite{wu2021cvt} for image classification in our cell Images}
    \label{CVT}
\end{figure}

\par
Among different variants of CvT, we employed CvT-13  as a base model consisting of 13 transformer blocks partitioning into multiple stages with a total of 19.98 parameters counts. Each stage contains two convolution based-operations i.e convolution token embedding and Convolution Projection. Convolution token embedding involves applying convolution operation on the input image with a fixed stride length having each patch overlapped with each other. This promotes spatial downsampling and at the sametime increasing the feature dimension of each token which is needed to capture the local information. Then, it employs convolution projection on 2D reshaped token maps where a depth-wise separable convolution operation is performed prior to self-attention network. The key and value projections are maintained to have the stride length of greater than 1 to ensure low computational cost. Finally the Multi-Head Self Attention(MHSA) operation is applied to the projected QKV allowing the network to fine-grained local context information. The whole process is repeated 3 times until a classification token is added at the end for image-recognition. The architecture is presented in Table \ref{cvt_table}. The author has utilized ImageNet-22k \cite{deng2009imagenet} to train their classification network. Taking the advantage of these weights, we modify the top level classifier and perform fine tuning without freezing the rest of the parameters for our targeted cell classes as shown in Figure \ref{CVT}.
\par
In the inference setting, we detect the outline of each segmented cell using a clustering algorithm from opencv library and utilize our classification model for further prediction. Additionally, we also count the total number of cells belonging to each cell type for cytology report generation.
\begin{table*}[h!]
\centering
\small 
\renewcommand{\arraystretch}{1.2} 
\setlength{\tabcolsep}{6pt} 

\adjustbox{max width=\textwidth}{ 
\begin{tabular}{|c|c|c|c|c|}
\hline
\textbf{Stage} & \textbf{Number of Conv. Transformer Block} & \textbf{Output Size} & \textbf{Layer Name} & \textbf{CvT-13} \\
\hline
\multirow{2}{*}{\textbf{Stage 1}} & \multirow{2}{*}{$N_1 = 1$} & \multirow{2}{*}{56 X 56} & Conv. Embed. & 7 × 7, 64, stride 4 \\
                                  &                            &                        & Conv. Proj.  & 3 × 3, 64 \\
                                  &                            &                        & MHSA         &  $H_1 = 1$, $D_1 = 64$ \\
                                  &                            &                        & MLP          & $R_1 = 4$ \\
\hline
\multirow{2}{*}{\textbf{Stage 2}} & \multirow{2}{*}{$N_2 = 2$} & \multirow{2}{*}{28 X 28} & Conv. Embed. & 7 × 7, 192, stride 2 \\
                                  &                            &                        & Conv. Proj.  & 3 × 3, 192 \\
                                  &                            &                        & MHSA         & $H_2 = 3$, $D_2 = 192$ \\
                                  &                            &                        & MLP          & $R_2 = 4$ \\
\hline
\multirow{2}{*}{\textbf{Stage 3}} & \multirow{2}{*}{$N_3 = 10$} & \multirow{2}{*}{14 X 14} & Conv. Embed. & 7 × 7, 384, stride 2 \\
                                  &                            &                        & Conv. Proj.  & 3 × 3, 384  \\
                                  &                            &                        & MHSA         & $H_3 = 1$, $D_3 = 384$ \\
                                  &                            &                        & MLP          & $R_3 = 4$ \\
\hline
\textbf{Head}                     &                            & 1 X 1                  & Linear       & 5 \\
\hline
\multicolumn{4}{|c|}{\textbf{Params}} & 19.61 M \\
\hline
\end{tabular}
}
\caption{CvT-13 model architecture. $H_i$ and $D_i$  refers to the number of heads and embedding feature dimension in the ith MHSA module. $R_i$ is the feature dimension expansion ratio in the ith MLP layer \cite{wu2021cvt}}
\label{cvt_table}
\end{table*}

\section{Experimental Results \& Discussions}
In this section, we discuss in brief about the datasets and evaluation metrics used for our study. Some implementation details are also provided for training and assessing our models. Subsequently, we evaluate and compare the performance of our proposed methodology for both segmentation and classification with some of the SOTA models. Various experiments are then conducted to show the robustness of our models both quantitatively and qualitatively.
\subsection{Datasets Used}
\subsubsection{Segmentation datasets}
\paragraph{\textbf{Cx22:}}
Cx22 is the first publicly available cervical cell segmentation dataset for various deep-learning applications \cite{liu2022cx22}. It masks 14,946 cellular instances in 1320 images by their ROIs-based label cropping algorithm. Among which, 500 image samples contain multi-cell instances and the rest of them have one pair of cervical cells that serve as supplements for other analysis. Each sample is annotated into corresponding cytoplasmic and nucleus regions as separate binary mask files. Since our main concern is to detect cells containing cytoplasm for cervical cancer diagnosis, we neglect annotation containing only the nucleus region. The training sample contains 400 images and the rest 100 are separated into test data with an image size of 512 by 512. But we only use 20 images with 751 ROIs for training purposes while the rest test samples are used for fair evaluation with other segmentation models.

\begin{figure}[t]
  \centering
  \begin{subfigure}{.375\columnwidth}
    \centering
    \includegraphics[width=0.8\linewidth]{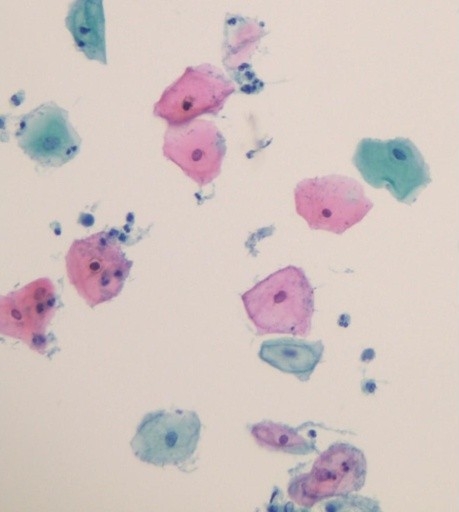}
  \end{subfigure}
  \hfill
  \begin{subfigure}{.55\columnwidth}
    \centering
    \includegraphics[width=0.8\linewidth]{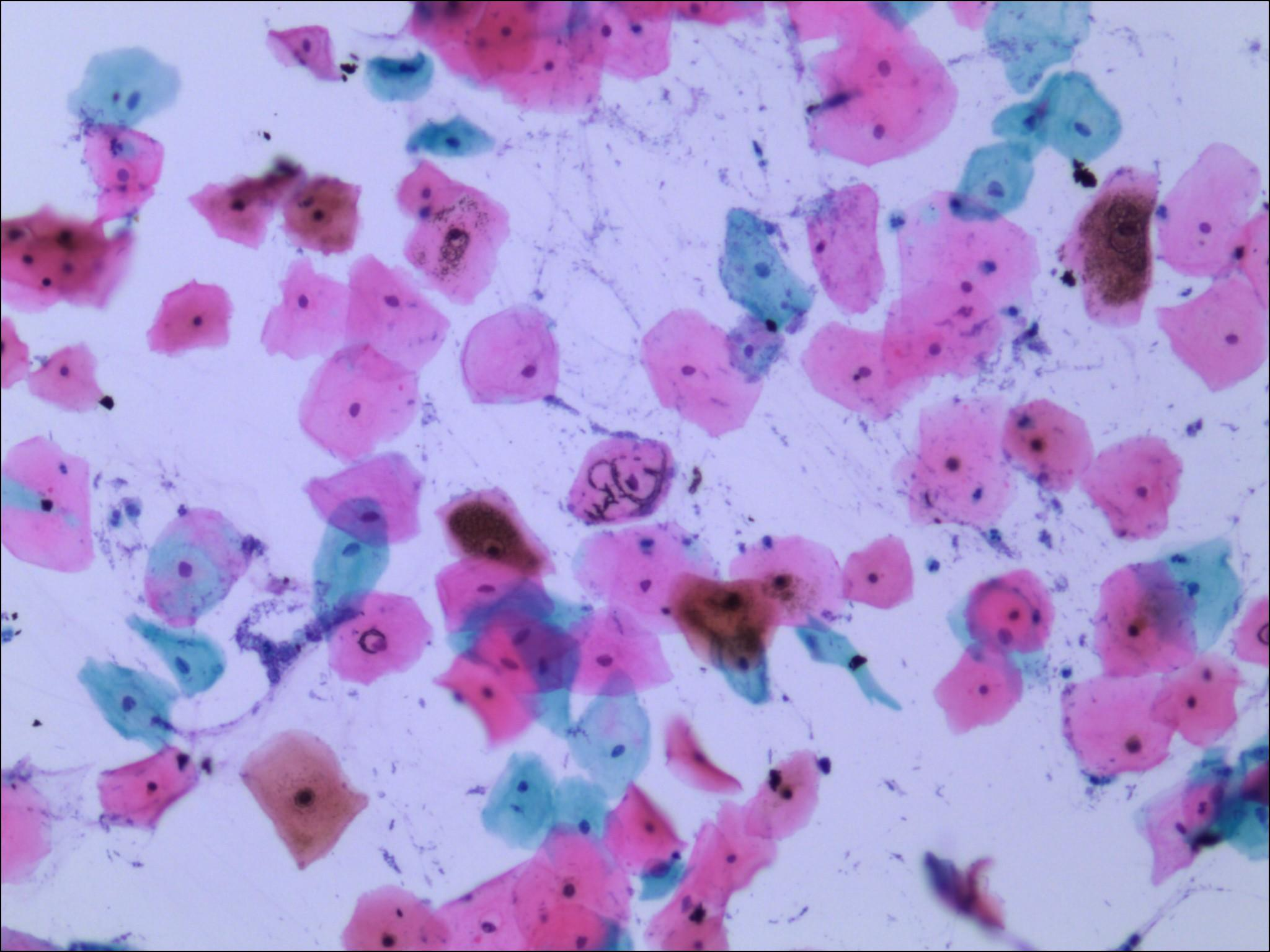}
  \end{subfigure}
  \caption{ Cytology Samples of Cx22 dataset \cite{liu2022cx22}}
\end{figure}

\paragraph{\textbf{CYTOCERVIX:}}
We manually collect the LBC samples using our automated microscope to build our cytology dataset. We collected approximately 2000 images from 10 different cytology slide samples by extracting non-overlapping image frames. Each slide differs from each other in terms of cell size, particle density, texture and color. Moreover, pathologists(beginner to experts) also have a significant impact on the preparation of LBC slide samples. So, learning from these heterogeneous dataset makes our model more robust to various unseen test samples. We only adopted 7 cytology images with a total of 1746 manually labeled ROIs for fine tuning our cellpose2.0 model using human-in-the-loop approach and the other 100 images are chosen for evaluation purpose with 10 images randomly sampled from each slide. The image size is 1500 by 1500 which is relatively larger in comparison to the Cx22. The annotation tools for test samples are developed by qualitatively evaluating the best performing segmentation model trained on our dataset with the experts. These test samples then serve as foundation for assessing various other segmentation models trained on other datasets.

\begin{figure}[t]
  \centering
  \begin{subfigure}{.45\columnwidth}
    \centering
    \includegraphics[width=0.8\linewidth]{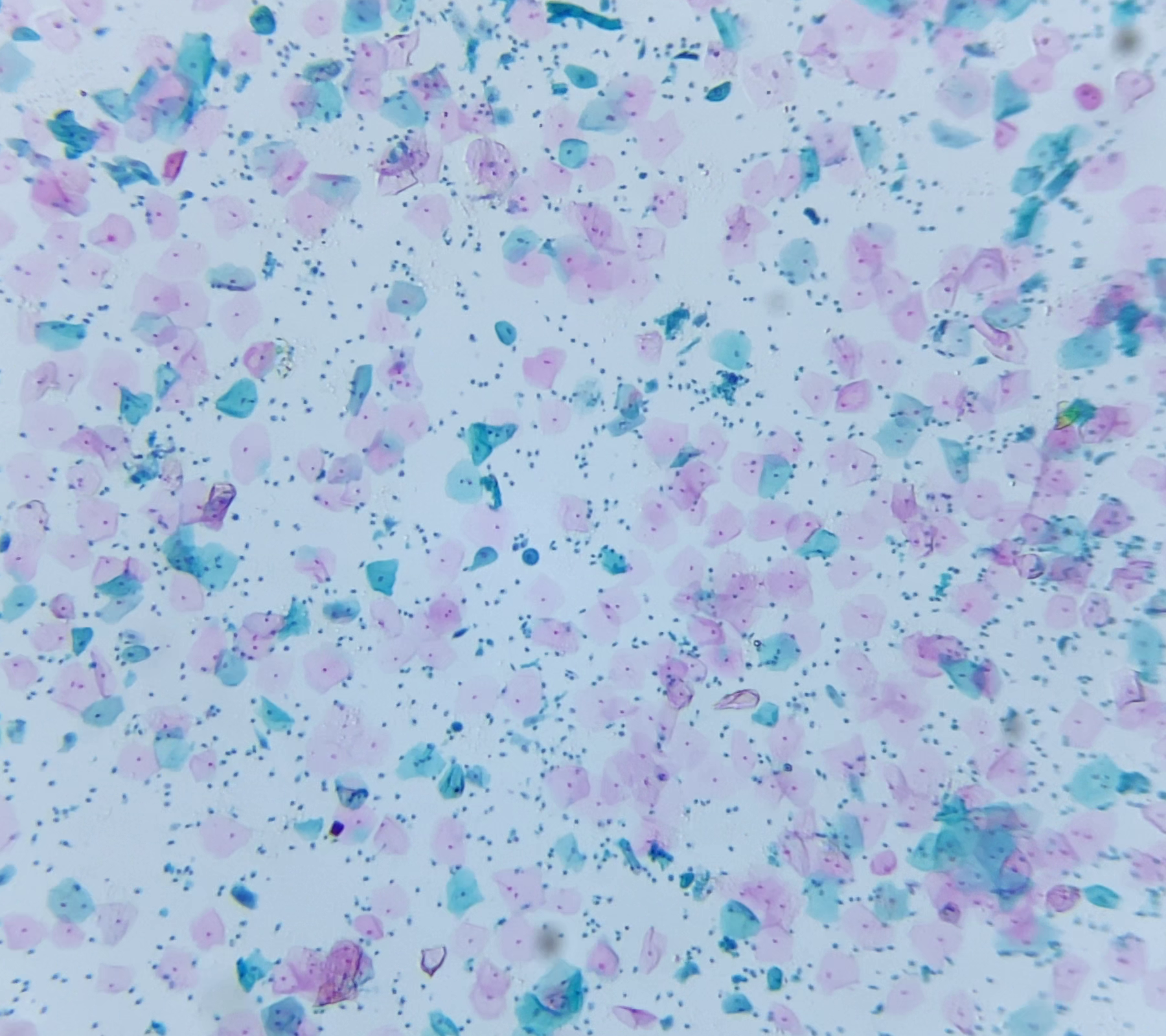}
  \end{subfigure}
  \hfill
  \begin{subfigure}{.40\columnwidth}
    \centering
    \includegraphics[width=0.8\linewidth]{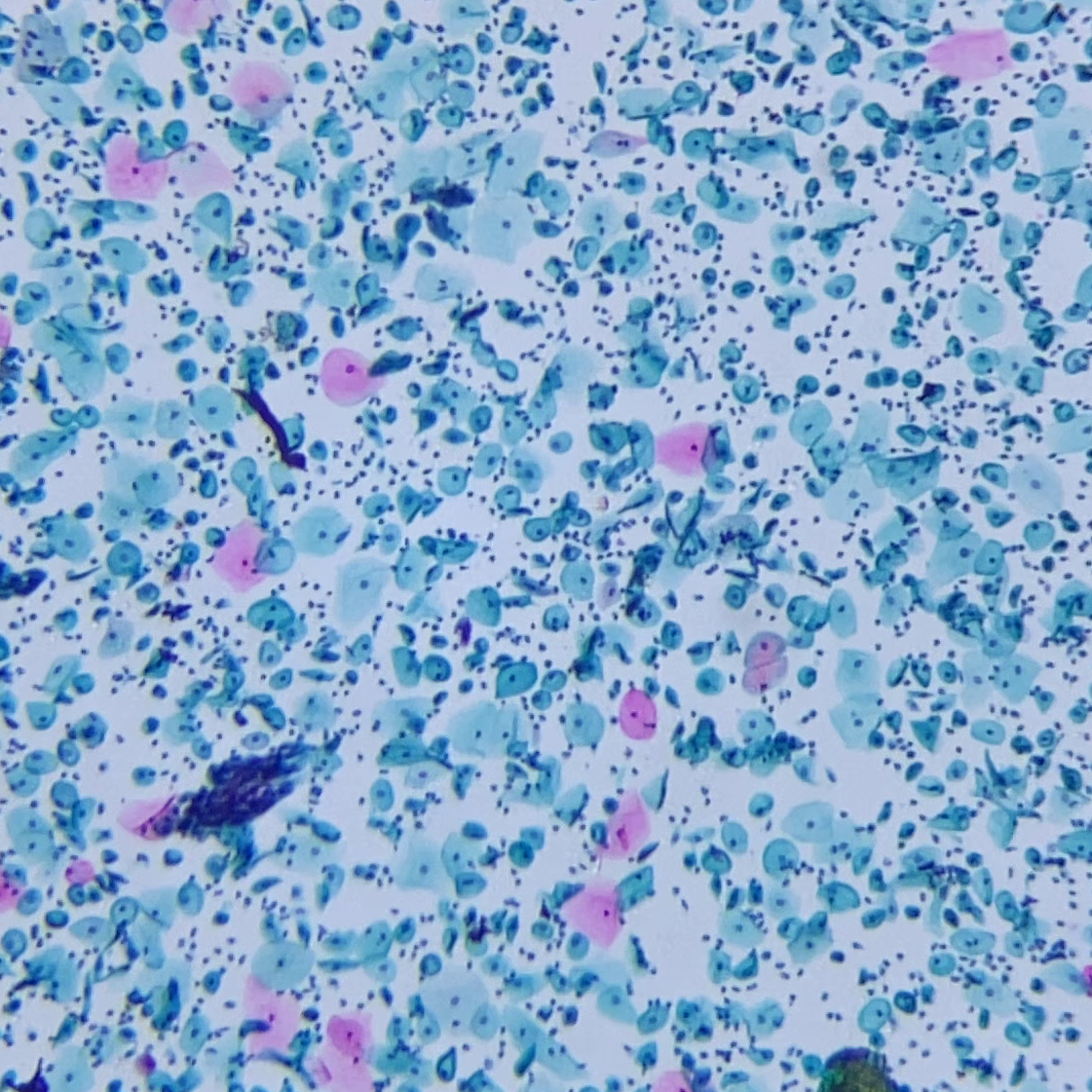}
  \end{subfigure}
  \caption{Cytology Samples of CYTOCERVIX}
\end{figure}

\subsubsection{Classification datasets}
\begin{figure}
\centering
\begin{subfigure}{.18\textwidth}
    \centering
    \includegraphics[width=.95\linewidth]{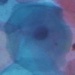}  
    \caption{Superficial-inter}
\end{subfigure}
\begin{subfigure}{.18\textwidth}
    \centering
    \includegraphics[width=.95\linewidth]{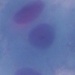}  
    \caption{Parabasal}
\end{subfigure}
\begin{subfigure}{.18\textwidth}
    \centering
    \includegraphics[width=.95\linewidth]{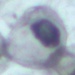}  
    \caption{Koilocytotic}
\end{subfigure}
\begin{subfigure}{.18\textwidth}
    \centering
    \includegraphics[width=.95\linewidth]{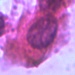}  
    \caption{Dyskeratotic}
\end{subfigure}
\begin{subfigure}{.18\textwidth}
    \centering
    \includegraphics[width=.95\linewidth]{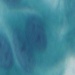}  
    \caption{Metaplastic}
\end{subfigure}
\caption{Cell samples of SIPaKMeD dataset \cite{plissiti2018sipakmed}}
\label{SIPaKMeD dataset}
\end{figure}

\paragraph{\textbf{SIPaKMeD dataset:}}
SIPaKMeD \cite{plissiti2018sipakmed} is the publicly available pap-smear datasets collected for the detection of cancerous and precancerous lesions cellwise based on their cellular appearance and morphology as shown in Figure \ref{SIPaKMeD dataset}. In total, it consists of 4049 annotated cell images into five different classes i.e (superficial-intermediate, parabasal) as normal, (koilocytotic, dyskeratotic) as abnormal and metaplastic as benign. Each cell size is resized to same dimension of 75 by 75 for training our classification model. Since each individual cell is separately classified, we assume the classification performance on our collected LBC samples at cell level doesn’t differ significantly from that of pap-smears. More detail of the dataset can be found in \cite{plissiti2018sipakmed}.

\subsection{Evaluation Metrics}
\subsubsection{Segmentation metrics}
We evaluate the performance of our segmentation model by comparing the degree of overlap between our predicted binary mask and ground truth segmentation. For these, all the mask cell instances above a certain threshold are labeled as positive and remaining background are treated as negative, reducing the whole problem as a binary segmentation. Sensitivity, Specificity and Dice similarity coefficient are the evaluation metrics used for our approach. \\
Sensitivity $= \dfrac{TP}{TP+FN}$, \\
Specificity $= \dfrac{TN}{TN+FP}$, \\
Dice coefficient $= \dfrac{2TP}{2TP + TN + FP}$ \\
These are the same metrics used for comparing our image segmentation performance with the existing approach.

\subsubsection{Classification metrics}
The performance of our 5-class classification model is evaluated by computing the commonly used classification metrics i.e average classification accuracy, precision, recall, F1 and  AUC score using 5-fold cross validation strategy. The dataset is divided into 5-folds, among which four are used for training and one fold is used for testing. Then, we average the results of all test folds. The confusion matrix and ROC curve are also plotted to give class-wise assessment for one of the test folds.

\subsection{Implementation details}
We used the pytorch framework for implementing our models(CvT and Cellpose2.0) for both training and testing purposes. Cellpose2.0 provides an efficient and lightweight GUI environment for implementing a human-in-the-loop approach for our target datasets. Users can upload their desired cytology images on one-click stored in the single folder. Then, existing pretrained models(e.g cyto2) are employed to generate the ROIs followed by human-correction. The masks images are saved in "\_seg.npy” format allowing the users to fine-tune their model at once when the annotation procedures are done. The model is trained for 120 epochs with the learning rate and weight decay of 0.05 and 0.00005 respectively. The cellpose2.0 requires us to specify two channels i.e each for cytoplasm  and nucleus. We found cytoplasm segmentation to be more recognizable at green channel with nucleus segmentation being found to be effective at grayscale. The cell diameter is calibrated automatically for different image samples. The flow threshold is set to 0.5. 
\par
For cell classification, we used Adam optimizer \cite{kingma2014adam} with the learning rate of 0.005 with ReduceLROnPlateau scheduler. The CvT model is trained for a total of 100 epochs with the batch size of 64 having the best model saved on low cross-entropy loss and higher classification accuracy evaluated on validation dataset. The entire setup is trained and evaluated on the Nvidia 1050 4GB GPU.

\subsection{Quantitative and Qualitative Evaluation}
\subsubsection{Segmentation}
We evaluate our model performance with various recent sota methods. In the context of binary segmentation of cytology cells, Cx22 \cite{liu2022cx22} proposes various baseline models based on variants of U-net trained on their own annotated dataset. Their best baseline obtained a Dice coefficient, sensitivity and specificity of 0.948, 0.954, 0.9823 respectively on test dataset. On our side, we employed four different “cellpose2.0” based segmentation models trained on the specific or combined cytology datasets. Our best performing model performs comparably with the above baseline model given the fact that the generalizability of Cx22 is still unknown for other unseen data.

Initially, the generalized “cellpose2.0” based cyto2 is assessed directly on both Cx22 and CYTOCERVIX samples. We found pretty good evaluation results on the Cx22 dataset as the model is able to capture the data distribution more accurately. It can be attributed to various factors such as clear distinction between ROIs and background, larger cell size, less or no background particles such as nucleus. In contrast, it performs poorly on CYTOCERVIX due to variation in cytology samples, cell size, objects densities and so on. We can qualitatively assess the results of the models in Table \ref{examples_segmentation}. The generalized “cellpose2.0” isn't able to segment most of the individual cells on the first sample collected using our microscope, though the result is pretty good for the last Cx22 sample. We fine tuned the next two variants of our models named Cx22\_cyto2 and CYTOCERVIX\_cyto2 separately on both Cx22 and CYTOCERVIX training datasets using a human-on-the-loop approach. Then, the performance of both models are evaluated on the Cx22 test samples. Cx22\_cyto2 provides very small difference in evaluation metrics of 0.76, 1.12 and 0.1 percentage compared to the generalized 'cyto2' model. 

However, the performance greatly increases by 53.72, 110.38 percent on both dice coefficient, sensitivity when using the CYTOCERVIX\_cyto2 model. This shows how the variation in data distribution would increase the performance on unseen datasets. The CYTOCERVIX\_cyto2 model is then qualitatively assessed by human-annotators based on the ability to distinguish each individual cell element and predicting binary segmentation on a few CYTOCERVIX test samples. This serves as the annotation tools for creating binary masks for our manually collected dataset for various deep learning tasks. We also compare the performance of the Cx22\_cyto2 model on our above annotated CYTOCERVIX samples. The proposed model generalized poorly on the unseen data. Lastly, we combined both datasets Cx22 and CYTOCERVIX for training another variant of “cellpose2.0” model named combined\_cyto2 and evaluated separately on their test sets. The results now show better generalization on both datasets despite an increased performance when trained on specific datasets. The performance is shown clearly in Table \ref{evaluate_seg_table}.
\begin{center} 
\begin{table*} 
\adjustbox{max width=\textwidth}{ 
\begin{tabular}{| c | c | c | c | c |} 
\hline 
 Cytology Samples & Generalized ‘Cyto2’ &  Cx22\_cyto2 & CYTOCERVIX\_cyto2 & combined\_cyto2 \\ 
\hline 
 \includegraphics[scale = 0.1]{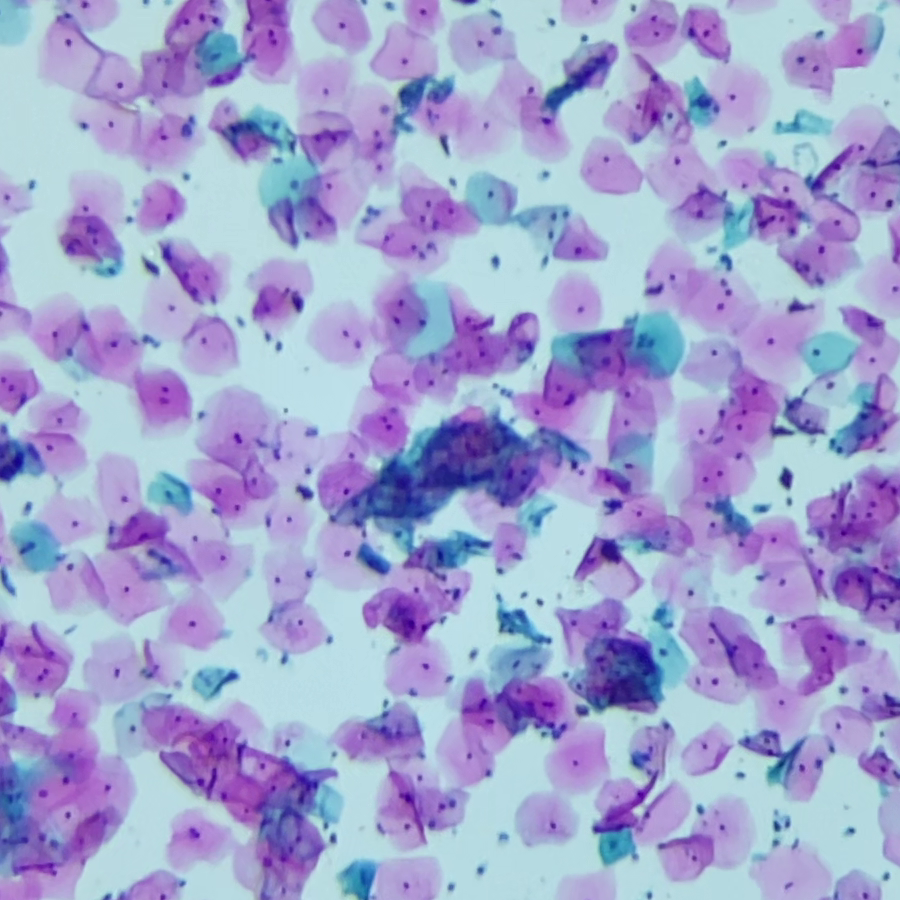} & \includegraphics[scale = 0.1]{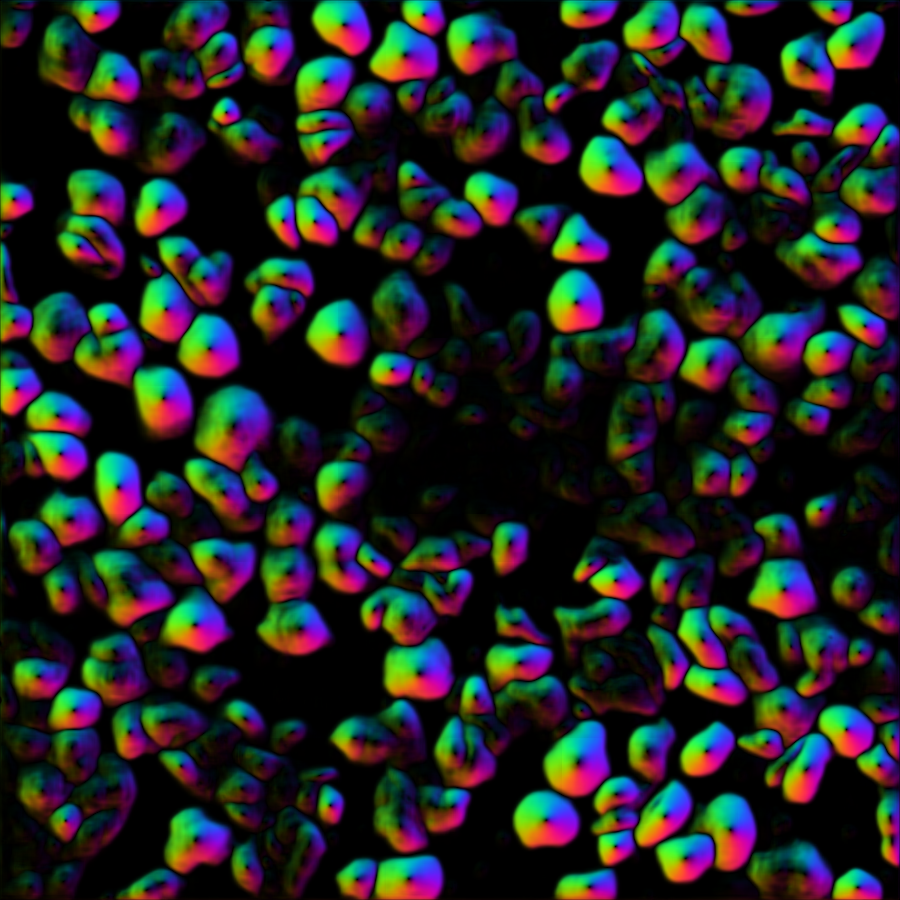}  & \includegraphics[scale = 0.1]{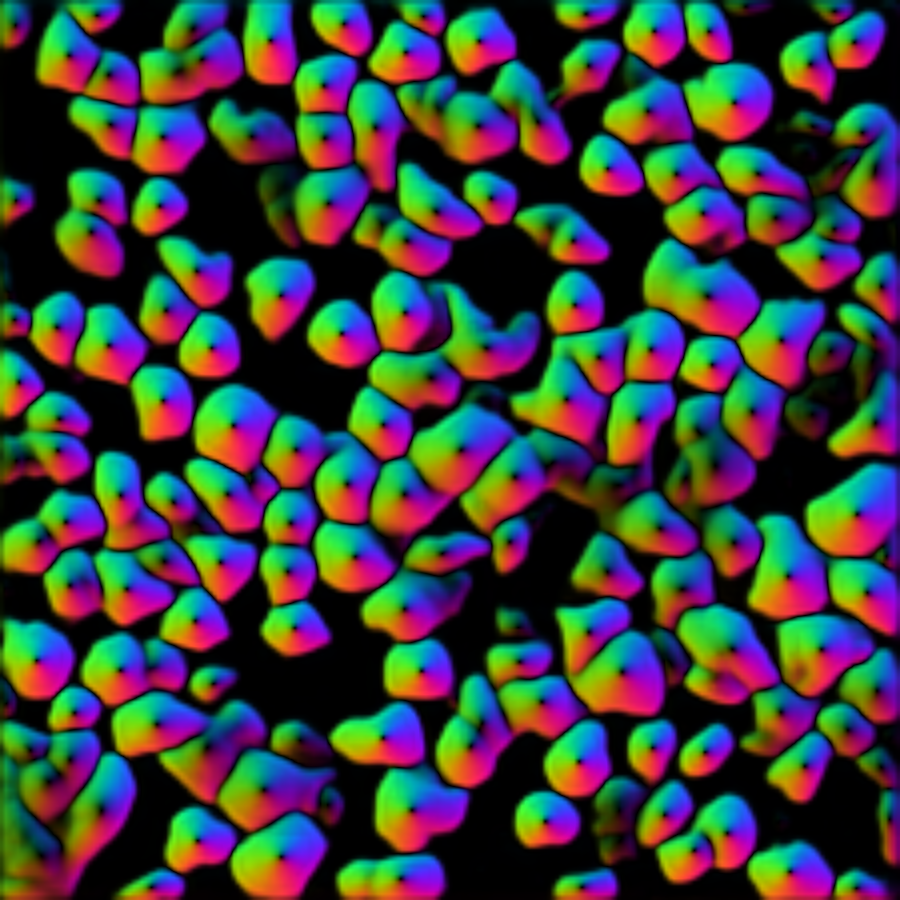}  & \includegraphics[scale = 0.1]{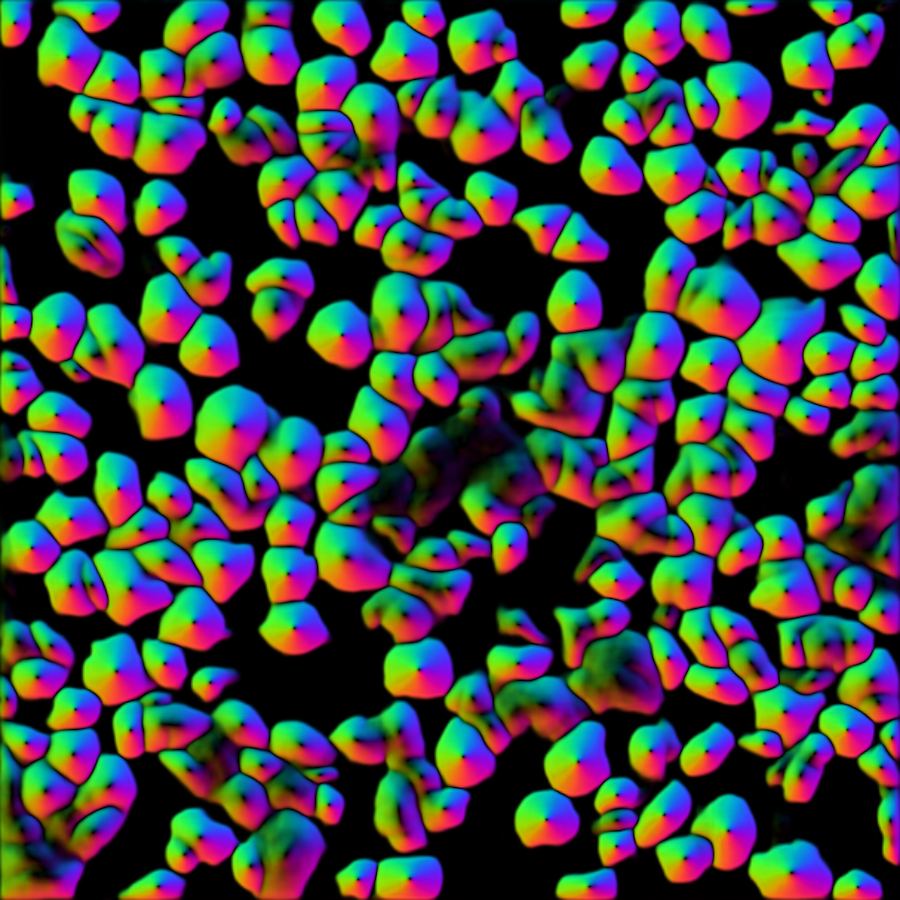}  & \includegraphics[scale = 0.1]{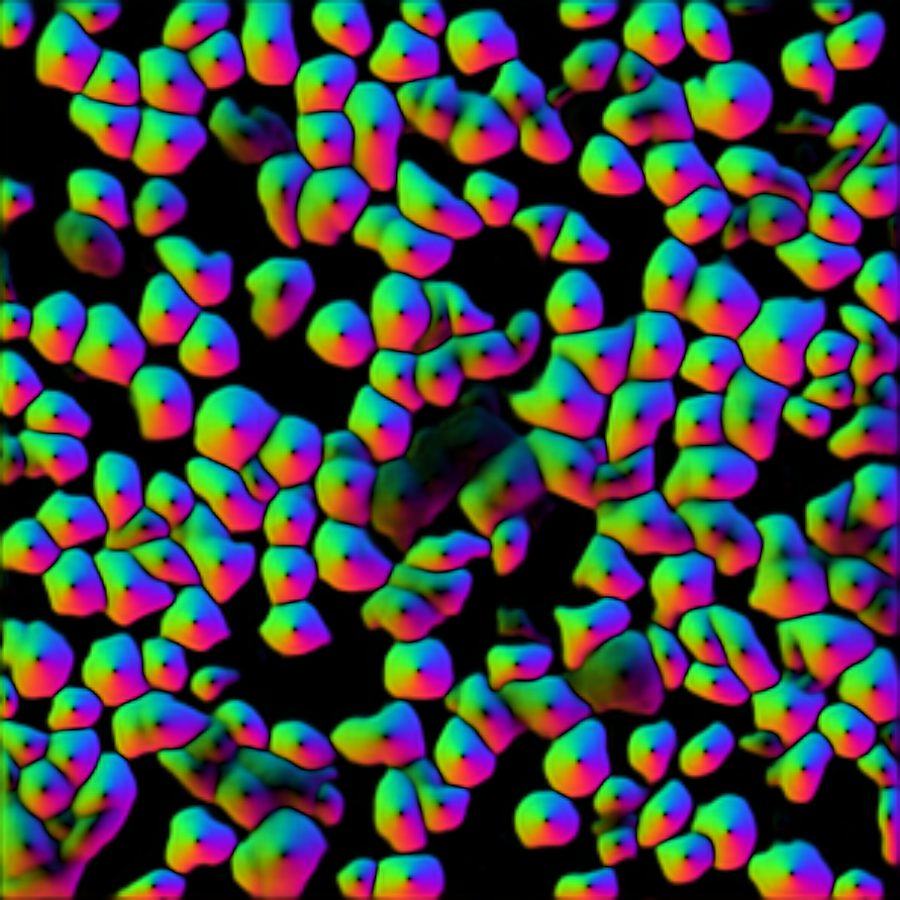} \\
 & \includegraphics[scale = 0.1]{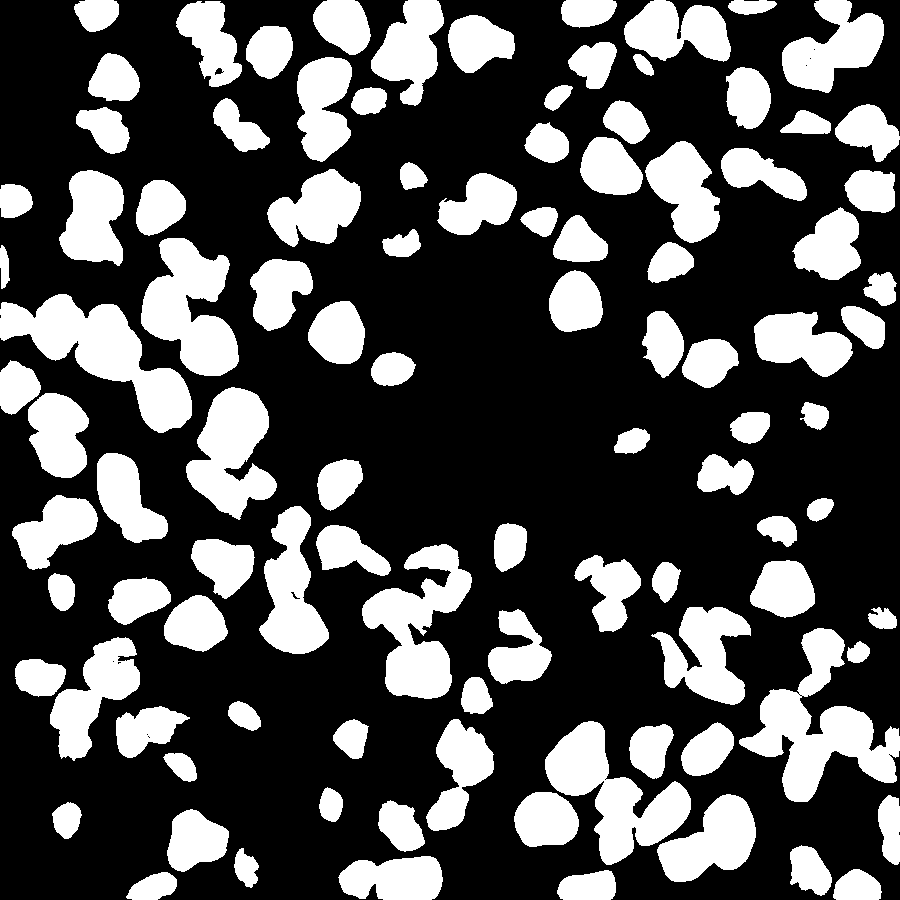}  & \includegraphics[scale = 0.1]{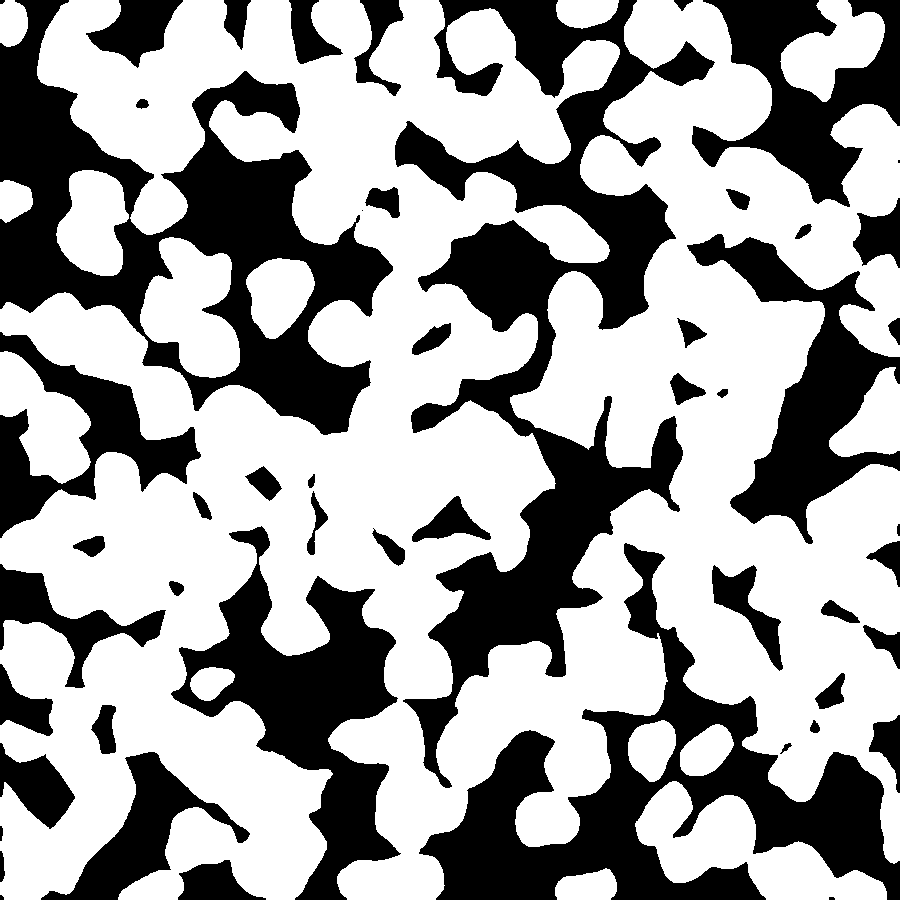}  & \includegraphics[scale = 0.1]{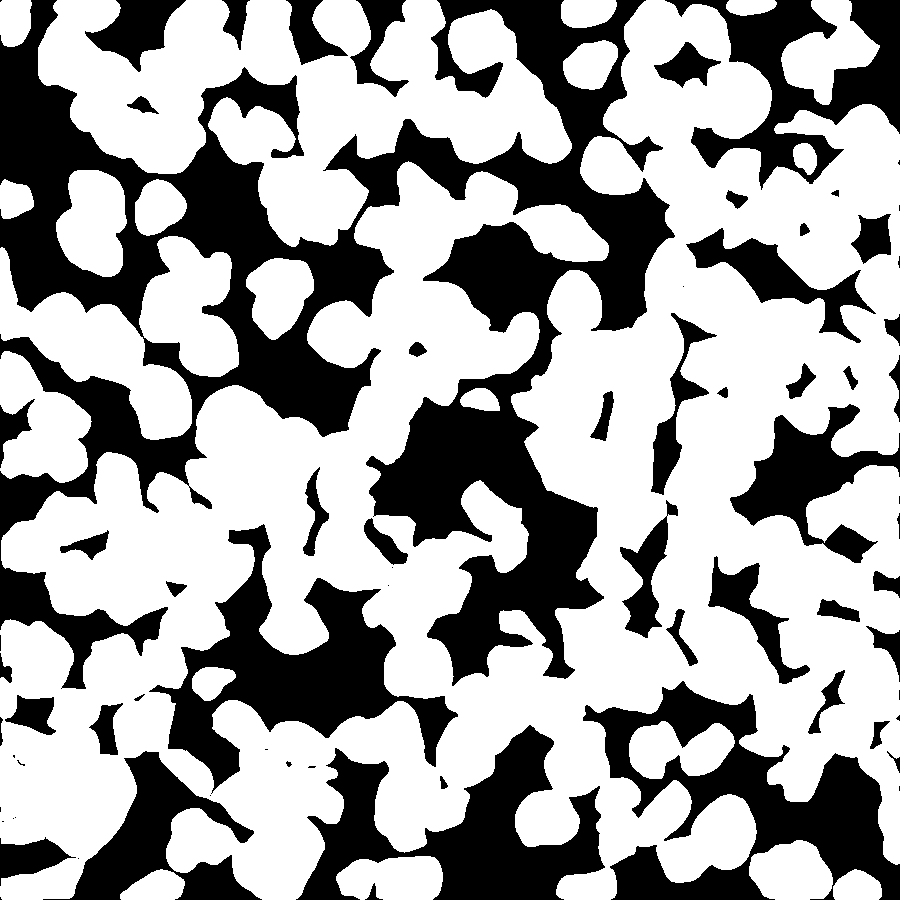}  & \includegraphics[scale = 0.1]{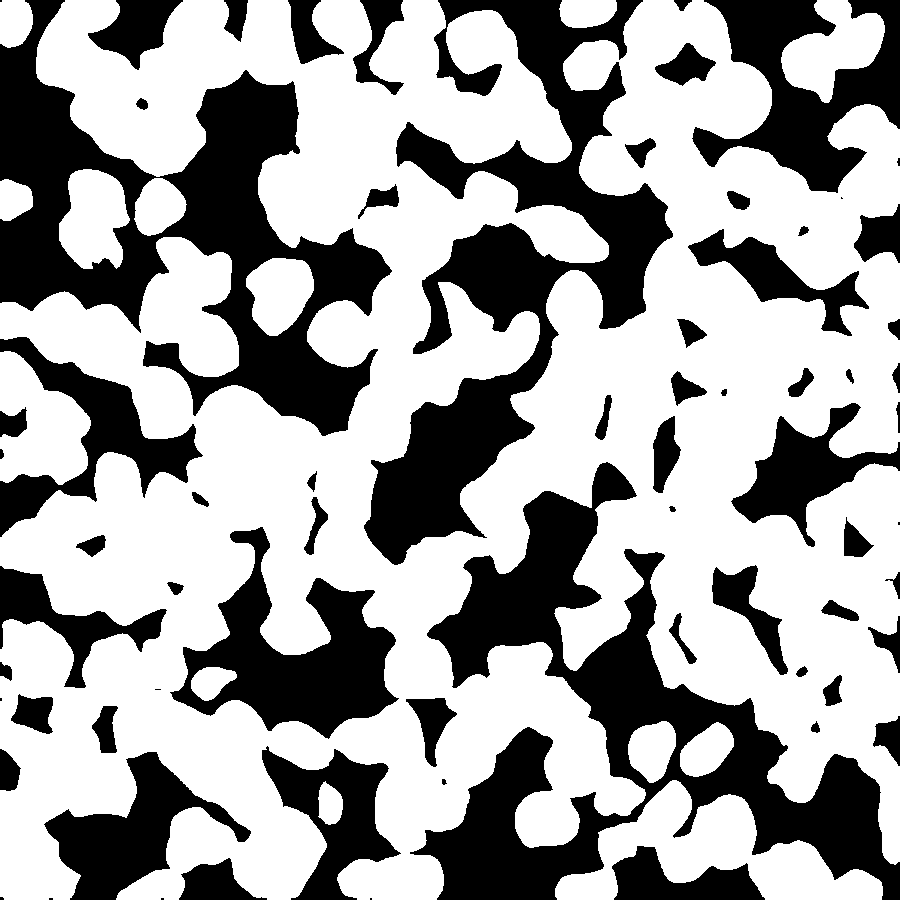}\\ 
 \hline 
\includegraphics[scale = 0.1]{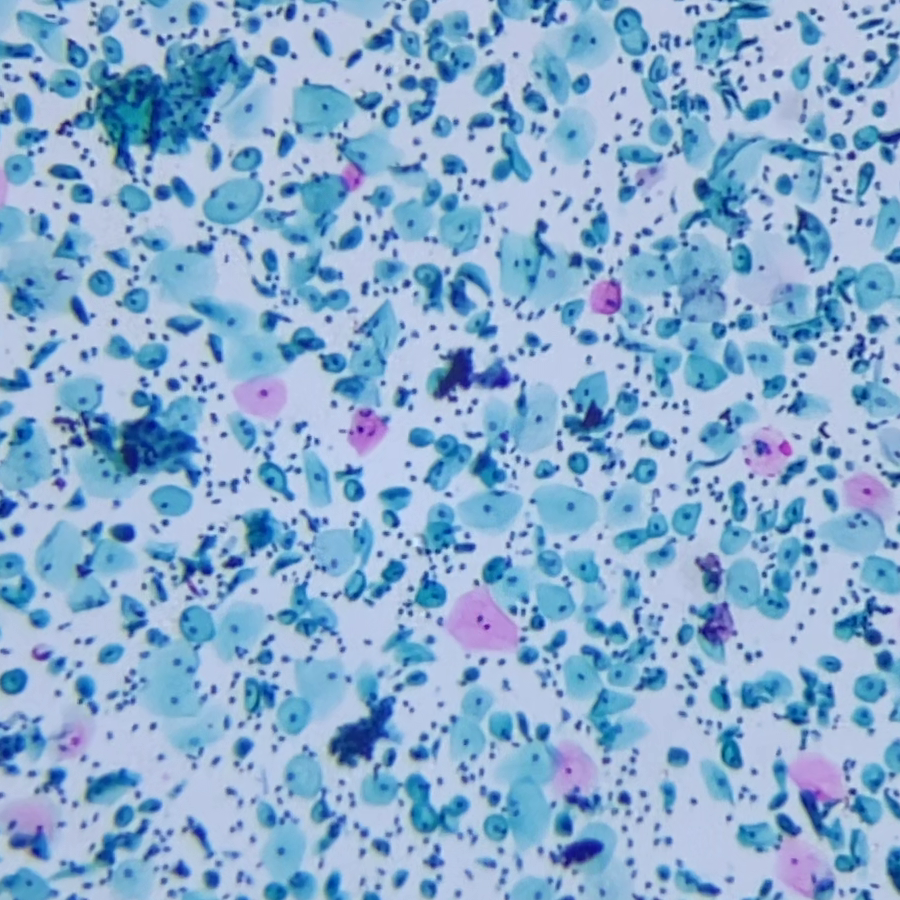} & \includegraphics[scale = 0.1]{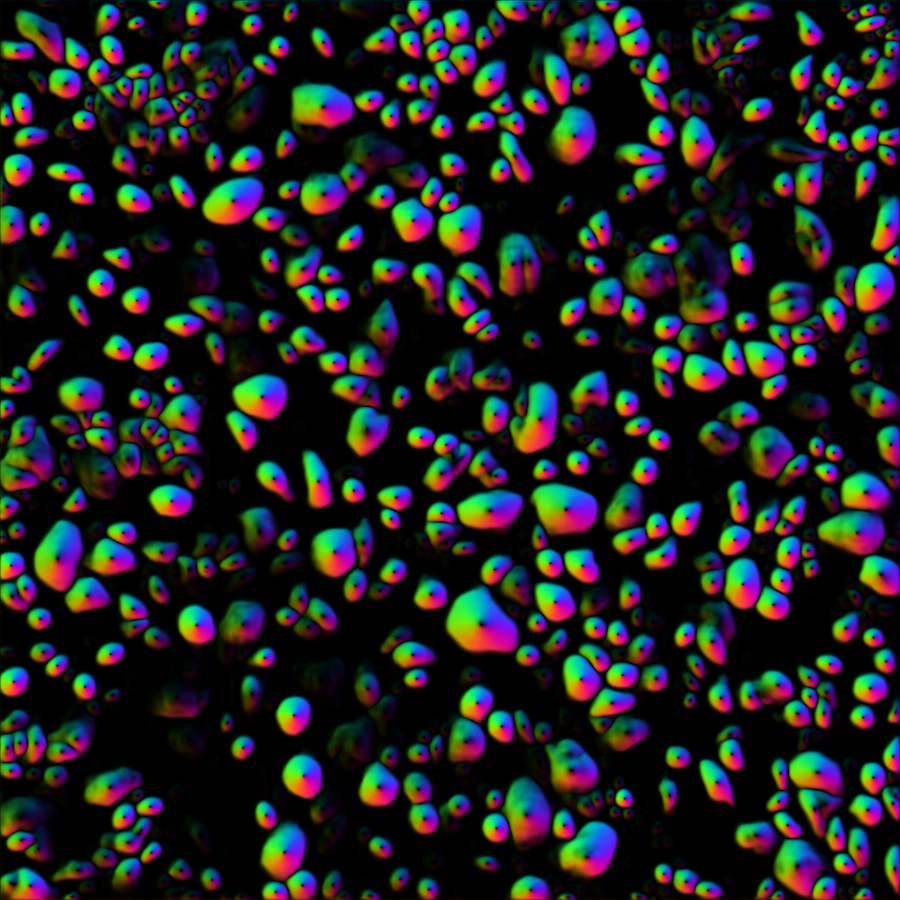}  & \includegraphics[scale = 0.1]{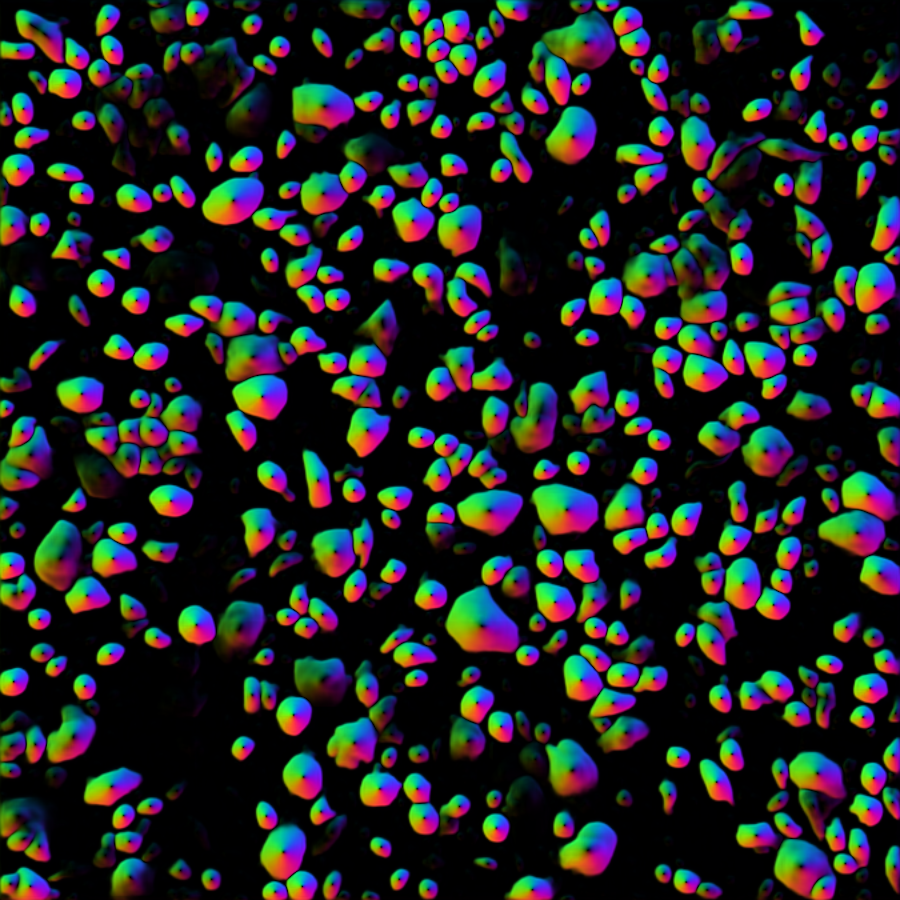}  & \includegraphics[scale = 0.1]{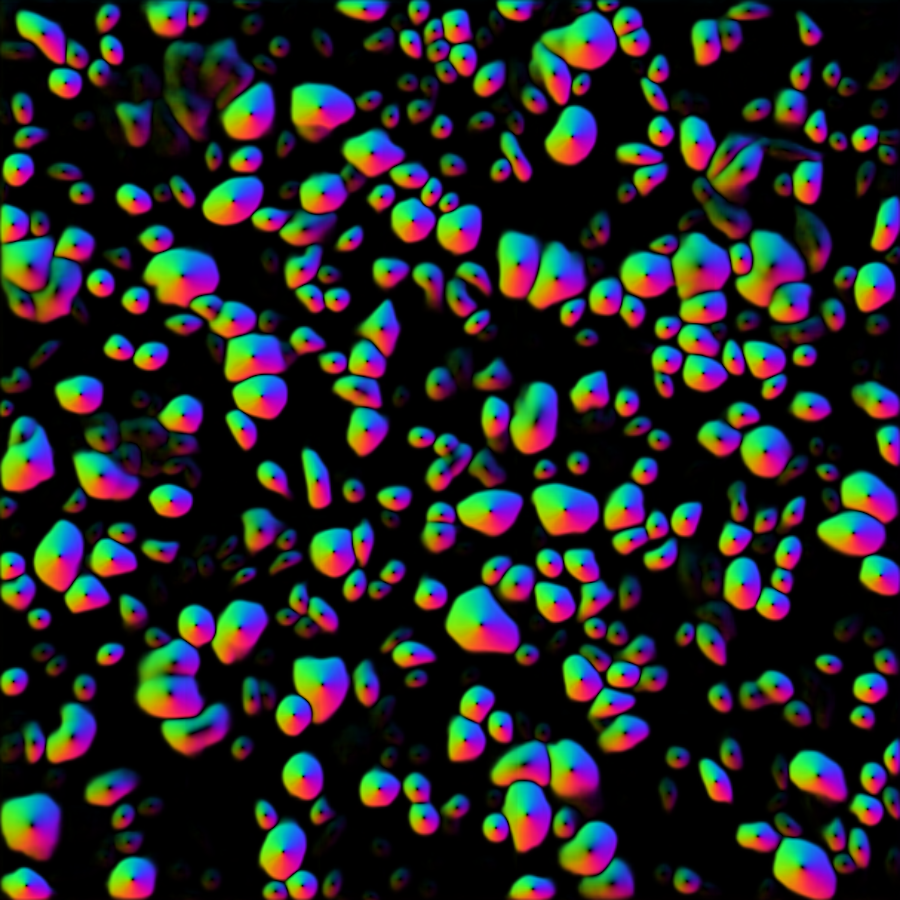}  & \includegraphics[scale = 0.1]{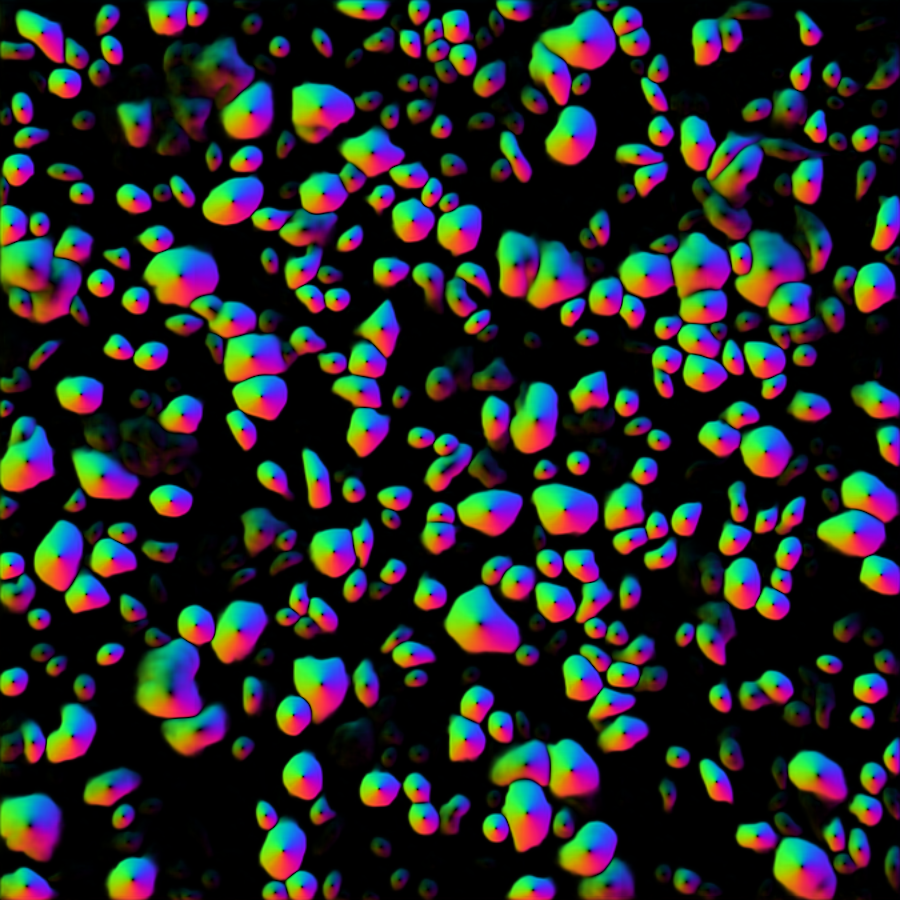}\\ 
 & \includegraphics[scale = 0.1]{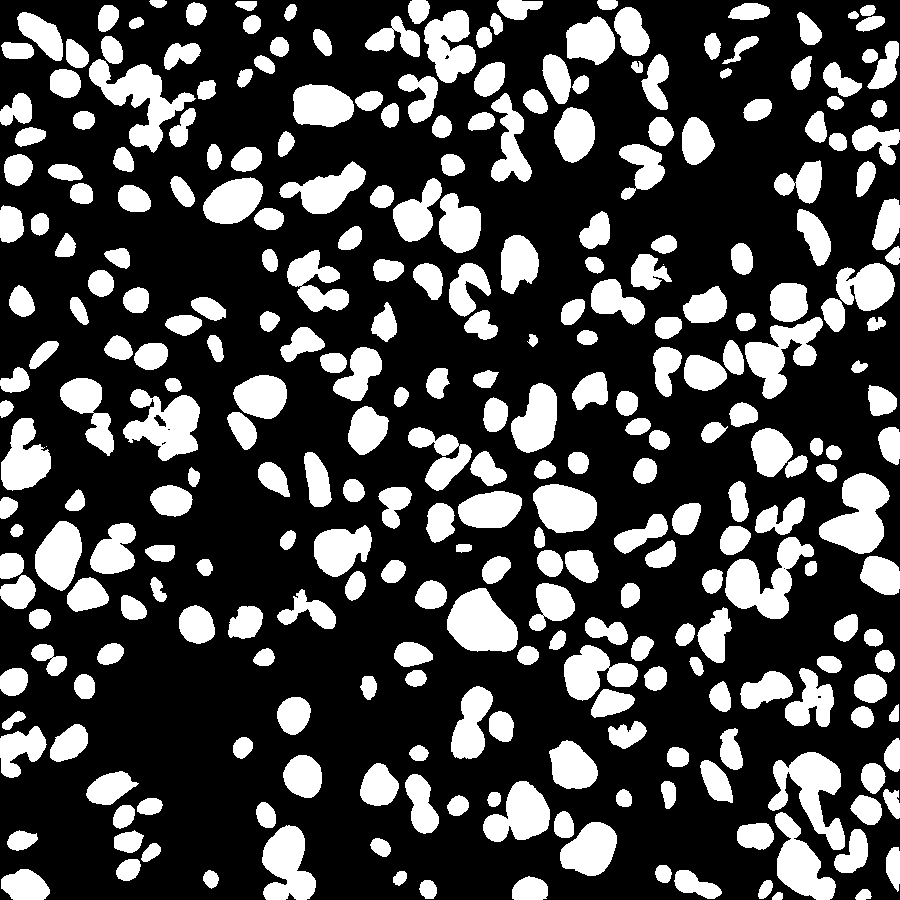}  & \includegraphics[scale = 0.1]{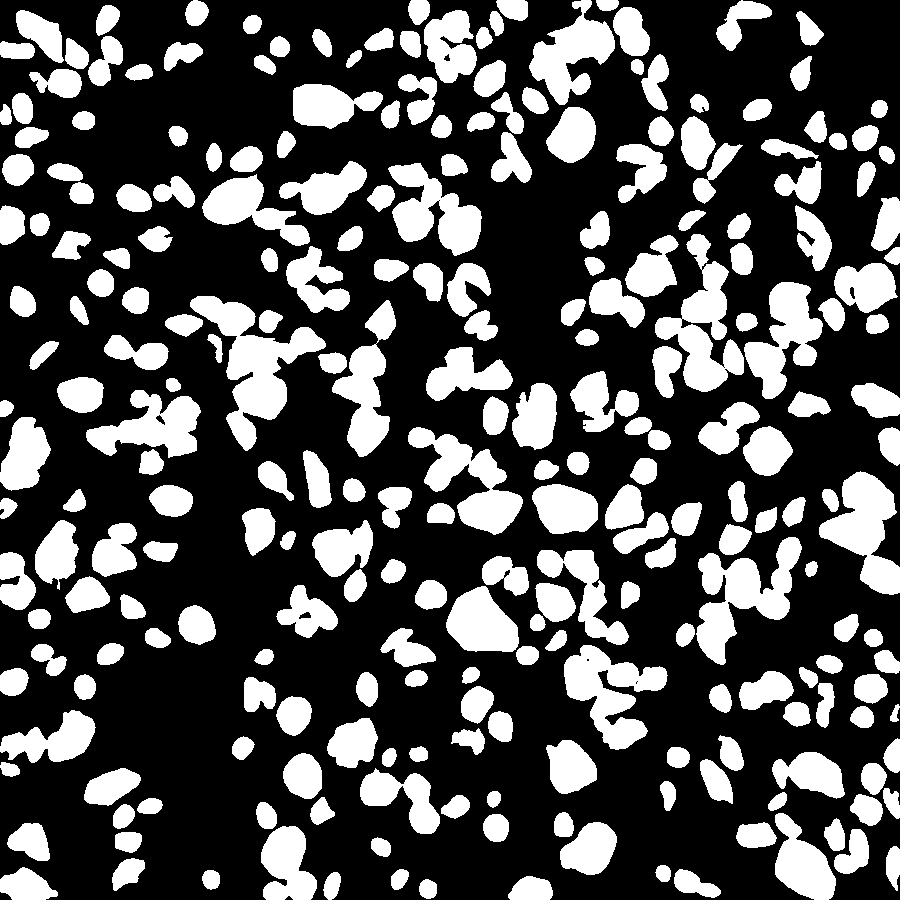}  & \includegraphics[scale = 0.1]{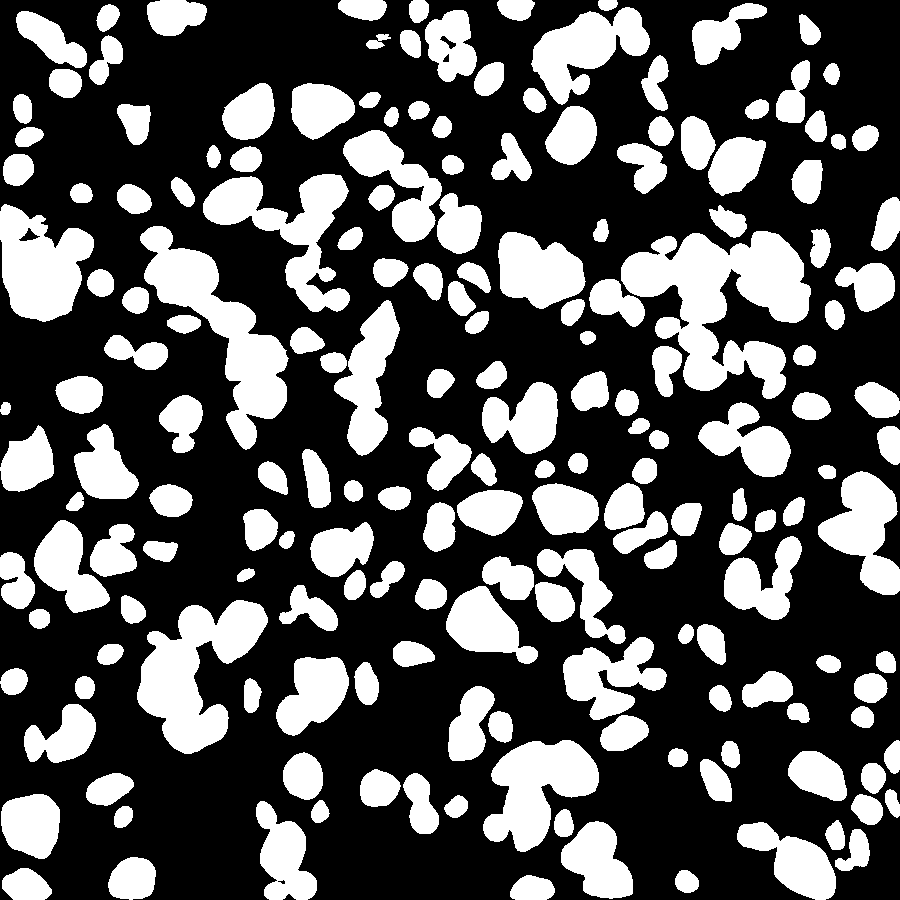}  & \includegraphics[scale = 0.1]{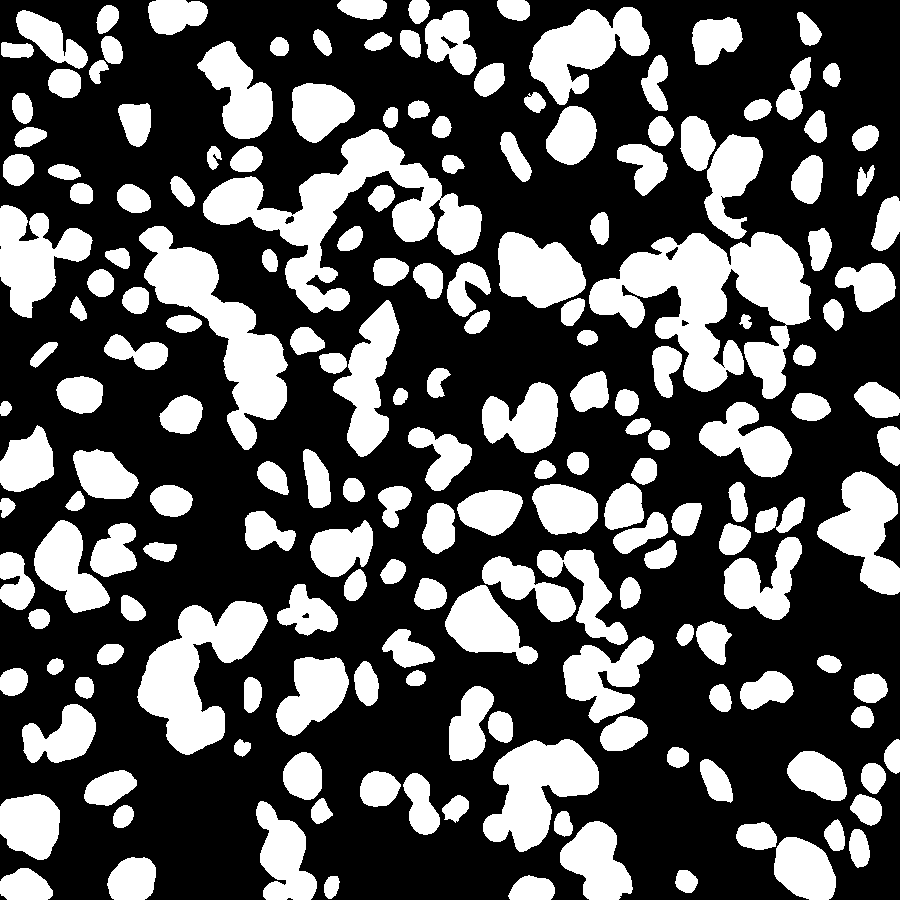}\\ 
 \hline 
\includegraphics[scale = 0.18]{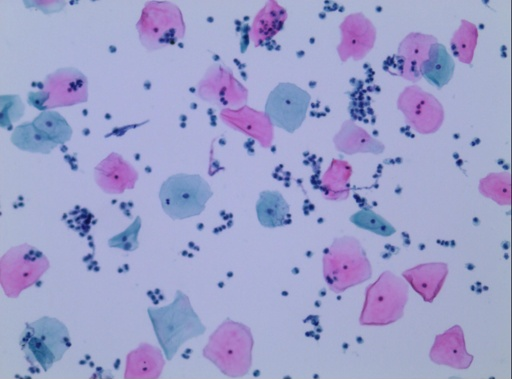} & \includegraphics[scale = 0.18]{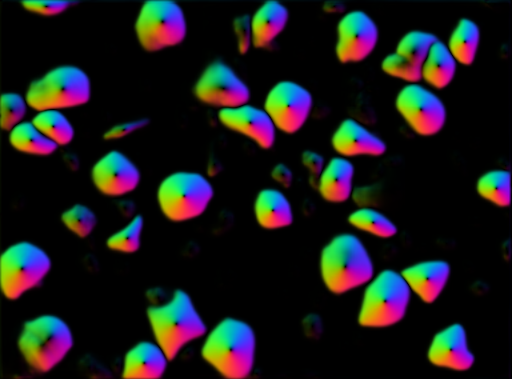}  & \includegraphics[scale = 0.18]{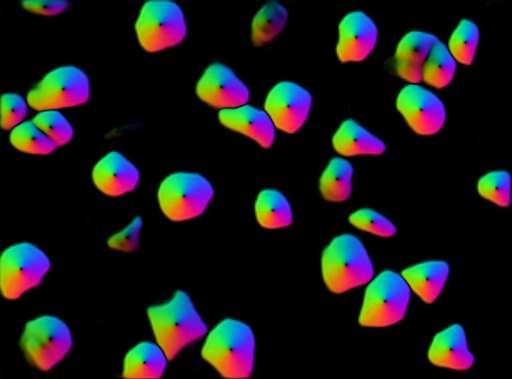}  & \includegraphics[scale = 0.18]{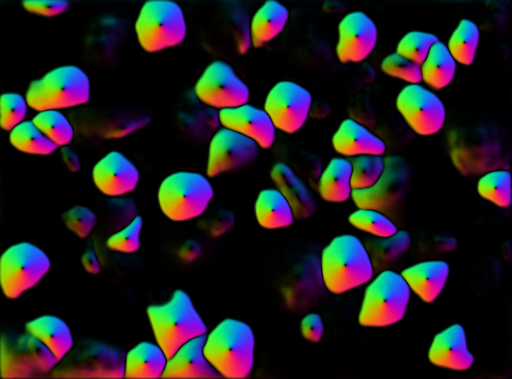}  & \includegraphics[scale = 0.18]{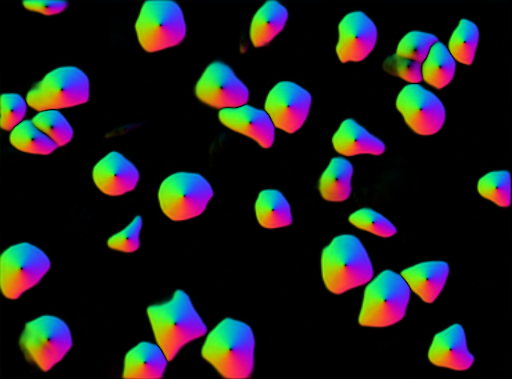}\\ 
 & \includegraphics[scale = 0.18]{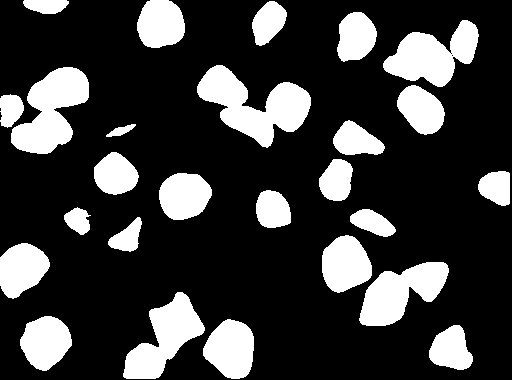}  & \includegraphics[scale = 0.18]{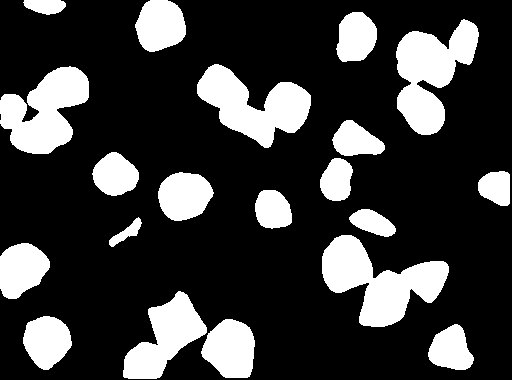}  & \includegraphics[scale = 0.18]{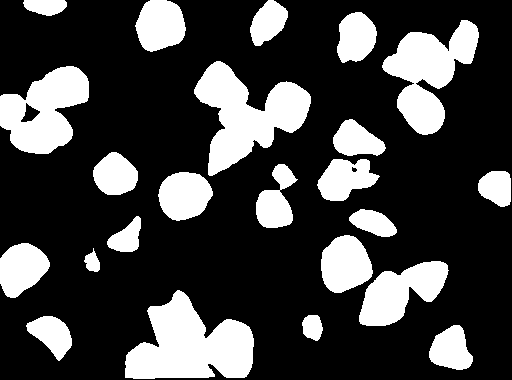}  & \includegraphics[scale = 0.18]{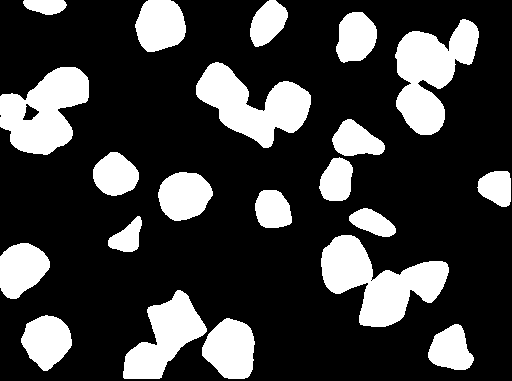}\\ 
 \hline
\end{tabular} 
}
\caption{Qualitative evaluation of cytology samples by predicting flows and binary masks from our variants of segmentation models. First two cytology images are the test samples from CYTOCERVIX and the last one is of Cx22 \cite{liu2022cx22}} 
\label{examples_segmentation} 
\end{table*} 
\end{center} 

\begin{table*}[!h]
\centering
\small 
\renewcommand{\arraystretch}{1.2} 
\setlength{\tabcolsep}{6pt} 
\adjustbox{max width=\textwidth}{ 
\begin{tabular}{|c|c c c|c c c|}
\hline
\multirow{2}{*}{\textbf{Methods}} & \multicolumn{3}{c|}{\textbf{Cx22}} & \multicolumn{3}{c|}{\textbf{CYTOCERVIX}} \\
\cline{2-7}
& \textbf{Dice coefficient} & \textbf{Sensitivity} & \textbf{Specificity} & \textbf{Dice coefficient} & \textbf{Sensitivity} & \textbf{Specificity} \\
\hline
\textbf{Baseline model} & \textbf{0.948} & 0.954 & \textbf{0.9823} & - & - & - \\
\hline
\textbf{Generalized ‘cyto2’} & 0.9322 & 0.9573 & 0.9715 & 0.6001 & 0.4556 & 0.9801 \\
\hline
\textbf{Cx22\_cyto2} & 0.9393 & 0.9689 & 0.9722 & 0.6988 & 0.6902 & 0.9367 \\
\hline
\textbf{CYTOCERVIX\_cyto2} & 0.9225 & 0.9585 & 0.9647 & - & - & - \\
\hline
\textbf{combined\_cyto2} & 0.9371 & \textbf{0.9805} & 0.9669 & \textbf{0.9164} & \textbf{0.9158} & \textbf{0.9816} \\
\hline
\end{tabular}
}
\caption{Segmentation evaluation results of our proposed models on both Cx22 and CYTOCERVIX.}
\label{evaluate_seg_table}
\end{table*}

 \begin{figure*}[h!]
    \centering
    \includegraphics[width=0.9\textwidth]{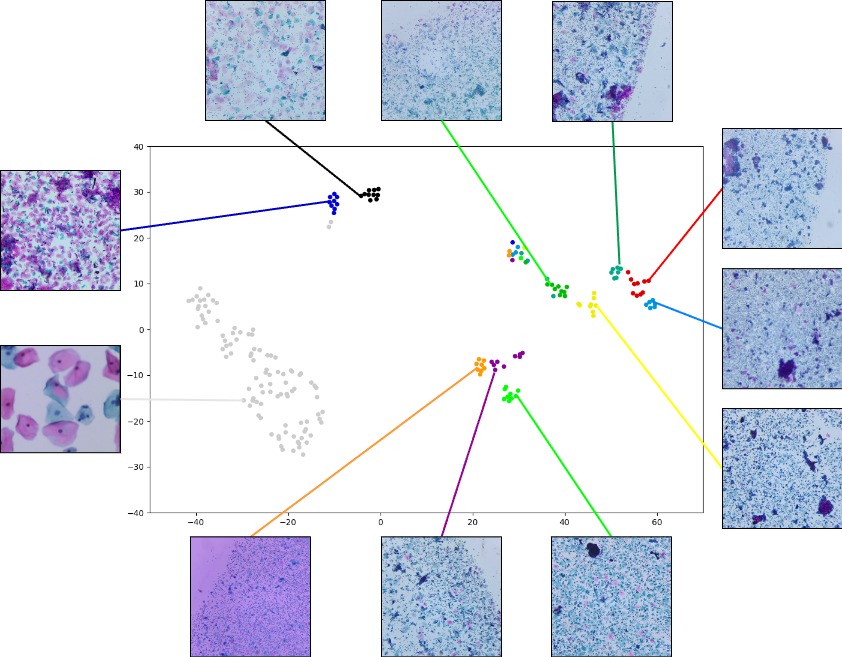}
    \caption{Visualization of diverse segmentation datasets using t-SNE embedding in 2-dimension. Each color represent different type of cytology samples. The point with grayscale represents the data from Cx22 while rest of points belongs to CYTOCERVIX dataset having 10 different types of images.}
    \label{tsne_embedding}
\end{figure*} 

 Moreover, we also assessed the variability of the segmentation datasets by applying t-SNE embedding to the image style vectors computed from combined\_cyto2 model. The style vectors are derived by using average global pooling to the lowest convolution maps of the network inspired from cellpose model. Sample images with similar characteristics are clustered together in the group while heterogeneous groups separate from each other as indicated by different colors in the Figure \ref{tsne_embedding}. The gray points which differs largely from our CYTOCERVIX in the embedding space indicate why model trained solely on Cx22 failed to generalize. However, we found two types of images i.e dark blue and black colored samples from other distribution with similar representation lies in the vicinity of some Cx22 samples contributing to given segmentation performance in Table \ref{evaluate_seg_table}. Much improvement can be seen other way when trained specific on our samples due to the considerable variation found in our collected CYTOCERVIX dataset which contribute overally not only for Cx22 dataset but also for other types of cytology samples we failed to collect at the present moment.

\subsubsection{Classification}
We used recent state-of-the art models to compare our classification performance on the 5-class SIPaKMeD dataset. Accuracy, precision, recall, f-score are taken as evaluation metrics to judge our results. The metrics are computed by averaging the results from a 5-fold validation strategy. 
We used two ways to train our model. The model is either trained from scratch or fine tuned with ImageNet-22k pretrained weights. The fine-tuning process requires less time with higher performance as shown in Table \ref{classification_table}. The results shows about approximately 1 percent gain in overall classification scores for SIPaKMeD dataset as compared to training from scratch.

\begin{table*}[h!]
\centering
\begin{tabular}{|p{5cm}|p{2cm}|p{2cm}|p{2cm}|p{2cm}|}
\hline
\textbf{Methods}  &   \textbf{Accuracy} & \textbf{Precision} & \textbf{Recall} & \textbf{F-score} \\ \hline
HDFF \cite{rahaman2021deepcervix}      & 99.14             & 99.20              & 99.00           & 99.00            \\ \hline
Fuzzy Rank \cite{manna2021fuzzy} & 95.43             & 95.34              & 95.38           & 95.36            \\ \hline
DeepCell-v2 \cite{fang2022deep} & 95.62             & 95.68              & 95.64           & 95.63            \\ \hline
Diff \cite{fang2024deep}         & 96.02             & 96.09              & 96.04           & 96.04            \\ \hline
MaxCerVixT \cite{pacal2024maxcervixt}   & 99.02             & 99.03              & 99.04           & 99.02            \\ \hline
CvT-13 from scratch (Ours)   & 98.76             & 98.77              & 98.79           & 98.77            \\ \hline
\textbf{CvT-13 finetuned (Ours)} & \textbf{99.68}    & \textbf{99.68}     & \textbf{99.67}  & \textbf{99.68}   \\ \hline
\end{tabular}
\caption{Performance comparison of our classification model (CvT-13) with various SOTA methods.}
\label{classification_table}
\end{table*}

Specifically, the fine-tuned CvT-13 model reached an accuracy of 99.68\%, which is a 0.54\% improvement over the HDFF model by Rahaman et al., which had previously held the highest reported accuracy of 99.14\%. Manna et al.'s Fuzzy Rank ensemble method and Fang et al.’s DeepCell-v2 achieve a similar score around 95.5, which is significantly lower than our model. Even if Fang et al. proposes the approach to synergistically integrate both CNN and transformer features in deep fashion to obtain a more robust feature for cell classification, their performance still fell short when compared to existing other classification models. On the other hand, MaxCerVixT is found to have better classification results by incorporating ConvNeXtBlockv2 block on MaxVit architecture. It surpasses our CvT-13 model learned from scratch but still outperformed by our fine-tuned model.

\begin{figure}[h!]
    \centering
    \includegraphics[width=7cm]{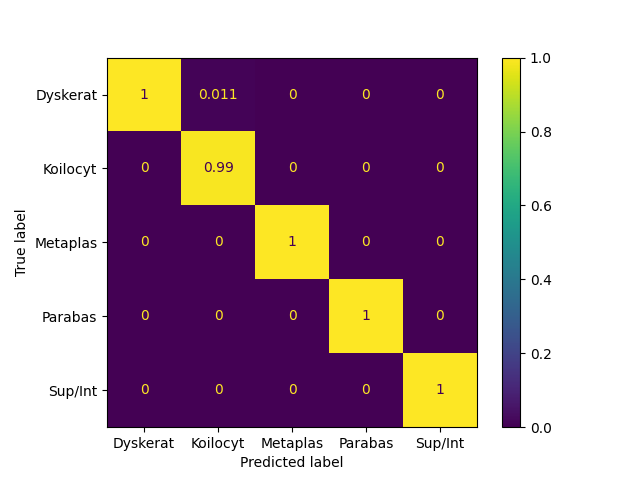}
    \caption{Confusion matrix for first fold validation on SIPaKMeD dataset}
    \label{confusion matrix}
\end{figure}

Moreover, we also plotted the confusion matrix for the first fold validation dataset consisting of 810 single cell images as shown in Figure \ref{confusion matrix}. The proposed model almost correctly classified all the cells which shows the strong classification ability. We also computed  the average AUC score of the proposed model and found it to be 0.9997. A high AUC score means that the model is generally reliable in identifying each cell type, reducing the chances of misclassification. These findings underscore the effectiveness of our approach in utilizing a pre-trained CvT-13 model for cervical cell classification, achieving not only higher accuracy but also more consistent performance across different evaluation metrics. This significant improvement highlights the potential of leveraging this simple and effective transfer learning techniques to enhance the performance of deep learning models in medical image classification tasks.

\section{Conclusion}
In this study, we proposed the comprehensive approach for analyzing cytology samples by integrating low-cost microscopes with our proposed AI methodology. The microscope is automated with the motorized system for efficient capturing of digital cytology samples collected from the maternity hospitals. The video is then processed through our image-stitching pipeline to get the full panoramic view of cytology images. The segmentation models are finetuned with human-in-the-loop approach requiring minimal ROIs. Lastly, CvT based classification model is proposed for cervical cytology images to classify a cell into 5-different classes. Each model is separately evaluated on different datasets which are either publicly available or collected through our mobile platform. The results comparison validates the state-of-the art performance of our proposed approach on a given cell-classification task. Furthermore, we also aim to make our collected segmentation dataset publicly available for further research.  

\bibliography{sample}


\end{document}